\documentclass[5p,times]{elsarticle}
\usepackage{graphicx} 
\usepackage{amsmath}
\usepackage[output-decimal-marker={,}]{siunitx}
\usepackage{subfig}
\usepackage{graphicx}
\usepackage{cleveref}
\usepackage{url}
\usepackage{algorithmic}
\usepackage{algorithm}
\usepackage{float}
\usepackage[english]{babel}
\usepackage{amsthm}
\usepackage{url}

\newtheorem*{remark}{Remark}

\begin{document}
\newcommand{\tp}{t^{\prime}}
\newcommand{\q}{\boldsymbol{q}}
\newcommand{\ub}{\boldsymbol{u}}
\newcommand{\p}{\boldsymbol{p}}
\newcommand{\dpb}{\boldsymbol{\dot{p}}}
\newcommand{\velbf}{\boldsymbol{v}}
\newcommand{\vel}{v}
\newcommand{\hvel}{\boldsymbol{\hat{v}}}
\newcommand{\dq}{\boldsymbol{\dot{q}}}
\newcommand{\dhq}{\boldsymbol{\dot{\hat{q}}}}
\newcommand{\ddq}{\boldsymbol{\ddot{q}}}
\newcommand{\taubf}{\boldsymbol{\tau}}
\newcommand{\x}{\boldsymbol{x}}
\newcommand{\Pb}{\boldsymbol{P}}
\newcommand{\R}{\mathbb{R}}
\newcommand{\Nat}{\mathbb{N}}
\newcommand{\y}{\boldsymbol{y}}

\newcommand{\cor}{\boldsymbol{c}}
\newcommand{\X}{\boldsymbol{X}}
\newcommand{\ab}{\boldsymbol{a}}
\newcommand{\thetab}{\boldsymbol{\theta}}
\newcommand{\Y}{\boldsymbol{Y}}
\newcommand{\w}{\boldsymbol{w}}
\newcommand{\Deltab}{\boldsymbol{\Delta}}
\newcommand{\deltab}{\boldsymbol{\delta}}
\newcommand{\Lag}{\mathcal{L}}
\newcommand{\E}{\mathbb{E}}
\newcommand{\norm}[1]{\left\lVert#1\right\rVert}
\newcommand{\D}{\mathcal{D}}
\newcommand{\f}{\mathbf{f}}
\newcommand{\e}{\boldsymbol{e}}
\newcommand{\mub}{\boldsymbol{\mu}}
\newcommand{\Fb}{\boldsymbol{F}}
\begin{frontmatter}



\title{Data efficient Robotic Object Throwing with Model-Based Reinforcement Learning}


\author[l1]{Niccolò Turcato}
\author[l1]{Giulio Giacomuzzo}
\author[l1]{Matteo Terreran}
\author[l1]{Davide Allegro}
\author[l1]{Ruggero Carli}
\author[l1]{Alberto Dalla Libera}
\affiliation[l1]{organization={Department of Information Engineering, Università degli Studi di Padova},
             addressline={Via Gradenigo 6/b},
             city={Padova},
             postcode={PD},
             country={Italy}}

\begin{abstract}
Pick-and-place (PnP) operations, featuring object grasping and trajectory planning, are fundamental in industrial robotics applications. Despite many advancements in the field, PnP is limited by workspace constraints, reducing flexibility. Pick-and-throw (PnT) is a promising alternative where the robot throws objects to target locations, leveraging extrinsic resources like gravity to improve efficiency and expand the workspace. However, PnT execution is complex, requiring precise coordination of high-speed movements and object dynamics. Solutions to the PnT problem are categorized into analytical and learning-based approaches. Analytical methods focus on system modeling and trajectory generation but are time-consuming and offer limited generalization. Learning-based solutions, in particular Model-Free Reinforcement Learning (MFRL), offer automation and adaptability but require extensive interaction time. This paper introduces a Model-Based Reinforcement Learning (MBRL) framework, MC-PILOT, which combines data-driven modeling with policy optimization for efficient and accurate PnT tasks. MC-PILOT accounts for model uncertainties and release errors, demonstrating superior performance in simulations and real-world tests with a Franka Emika Panda manipulator. The proposed approach generalizes rapidly to new targets, offering advantages over analytical and Model-Free methods.

\end{abstract}



\end{frontmatter}

\section{Introduction}
\label{sec:intro}
Pick-and-place operations are fundamental in robot manipulators' industrial applications. In a pick-and-place (PnP) operation, the robot manipulator grasps the object and plans a trajectory that brings it to the target position, also considering motion constraints due to the environment. Several works analyzed this problem and proposed different solutions, see, for instance, \cite{pnp, 5342467}.
Despite these advances, there still are situations where PnP is limited or cumbersome to implement: target locations must be inside the reachable workspace of the robot manipulator, thus limiting the flexibility in the work cell organization.  

Pick-and-throw (PnT) is a relatively unexplored alternative to PnP that shows great promise. In PnT,  the robotic arm throws the object to the target location instead of placing it there. This approach falls into the "extrinsic dexterity" applications, where robots exploit external resources (gravity and contacts) to move an object, see, for instance, \cite{extrinsic_dexterity, pmlr-v164-chen22a} in the context of manipulation. Throwing objects instead of placing them can significantly improve time efficiency and expand the manipulator's reachable workspace. 
Nonetheless, executing a PnT task in a real-life scenario presents considerable difficulties. Indeed, to perform a precise throw, the robot needs to take into account the object's characteristics, such as mass, shape, and inertia, as well as factors like friction, collisions, and discontinuous dynamics. Moreover, PnT entails high-frequency and high-speed movements while also requiring accurate coordination between the arm motion and the object release. However, a precise characterization of the object to throw and the phenomena that affect its dynamics are not always available, and, in general, they are hard to derive. Furthermore, several aspects, like communication nonidealities, can compromise the coordination between the arm motion and the object release. At high speeds, even small delays have significant impacts on the outcome. As a result, this is a relatively complex task even for humans, who, contrary to robots, stand out for their remarkable perception and coordination capabilities. 

Solutions proposed to solve the PnT problem can be grouped into analytical and learning-based. Analytical solutions consist of two steps. First, they characterize the system properties, then they focus on the trajectory generation. 
An example of PnT application in industry has been presented in \cite{urban_waste_throwing}, where a delta robot is integrated into a waste sorting system, with a throwing approach specifically tailored to the system.

Most of the times, analytical methods are tricky to implement. Indeed, fine-tuning procedures are necessary to avoid too crude approximations of the actual system that could compromise the algorithm's effectiveness. These operations are particularly time-consuming and must be performed every time the setup changes, thus limiting generalization. 
Learning-based solutions are an interesting alternative to the analytical approach. These solutions have the potential to learn an algorithm for the PnT task in an automated fashion, just from observations collected by interacting with the system. Most of the strategies proposed in the literature rely on Model-Free solutions, such as general-purpose Deep Networks \cite{lecun2015deep_learning} or Model-Free Reinforcement Learning (MFRL) \cite{ccalicsir2019mfrl_survey}, namely, algorithms that directly optimize the policy from data, without deriving a model of the system. Several works focused on the object-throwing task. 
Among learning-based approaches, an important class considers a setup where the end-effector is a cup or uses the gripper as a cup, see, for instance \cite{RL_policy_training_ball_throwing, hierarchical_RL_ball_throwing, kober2012reinforcement_learning_primitives, gutzeit2018besman_ball_throwing, robot_skill_learning_deep_autoencoder_ball_throwing}. 
Another group of works presents object-throwing strategies for robotic systems, using the gripper to grasp the object and to release while the arm is moving \cite{ball_throwing_visual_feedbacK_gripper, learning_ball_throwing_decision_transformers_gripper,huang2023dynamic_handover_gripper,tossingbot,pick_and_throw_sorting_deep_rl}. 
It is worth mentioning that \cite{tossingbot,pick_and_throw_sorting_deep_rl} also considered the grasping problem. 
\Cref{sec:related_work} reviews the cited contributions.

Collected results show the potentialities of learning PnT from data. Such approaches proved effective in solving the tasks, but typically require significant interaction time with the system to converge to good solutions, or do not generalize to arbitrary targets. In this work, with the intent to obtain good generalization performance with a limited amount of experience, we propose a Model-Based Reinforcement Learning (MBRL) solution. Different from MFRL, MBRL algorithms exploit data also to obtain a model of the system. Then, they use the model to simulate the system and limit the time spent on the actual system, see \cite{moerland2023mbrl_survey} for an overview.

This paper presents a learning-based framework to solve PnT tasks. First, we introduce Monte Carlo - Probabilistic Inference for Learning Object-Throwing (MC-PILOT), a MBRL solution derived from MC-PILCO \cite{mcpilco_tro}. Examples of applications of MC-PILCO are \cite{amadio2023mcpilco_raw_meas,wiebe2024reinforcement}.
\\Following the MBRL strategy, MC-PILOT exploits data collected on the actual system to derive a stochastic model of the object dynamics. The algorithm optimizes a policy for tossing the object to arbitrary locations, taking into account model uncertainties and release errors. Secondly, we present a procedure to estimate the release error distribution, in a setup where the robotic arm and the gripper are not synchronized in real-time; this is a case of interest for several real-life scenarios. Our solution exploits the data-driven model of the object dynamics, thus demonstrating an additional advantage of the Model-based approach. 

We carried out extensive tests in a simulated environment and also on an actual test-case scenario, implemented with a Franka Emika Panda manipulator. We compared our approach with both an analytic solution and a Model-Free approach, considering different object and target characteristics. Results show that the PnT can be a valid, robust and flexible strategy to extend the manipulator workspace and demonstrate the advantages of the proposed solution. Compared to Model-Free solutions, the algorithm (i) generalizes more rapidly to unseen target locations, and (ii) adapts rapidly to object and target variations. Interestingly, we show that MC-PILOT does not require re-exploration of the environment when the task requirements change, e.g. when the target height is modified. Instead, compared to analytical solutions, the data-driven approach can lead to more accurate solutions, without requiring time-expensive engineering phases. A demonstration video is available\footnote{Demo video: \url{https://youtu.be/0e8IWstunsc}}.

\begin{figure}
    \centering
    \subfloat{\includegraphics[width=0.95\columnwidth]{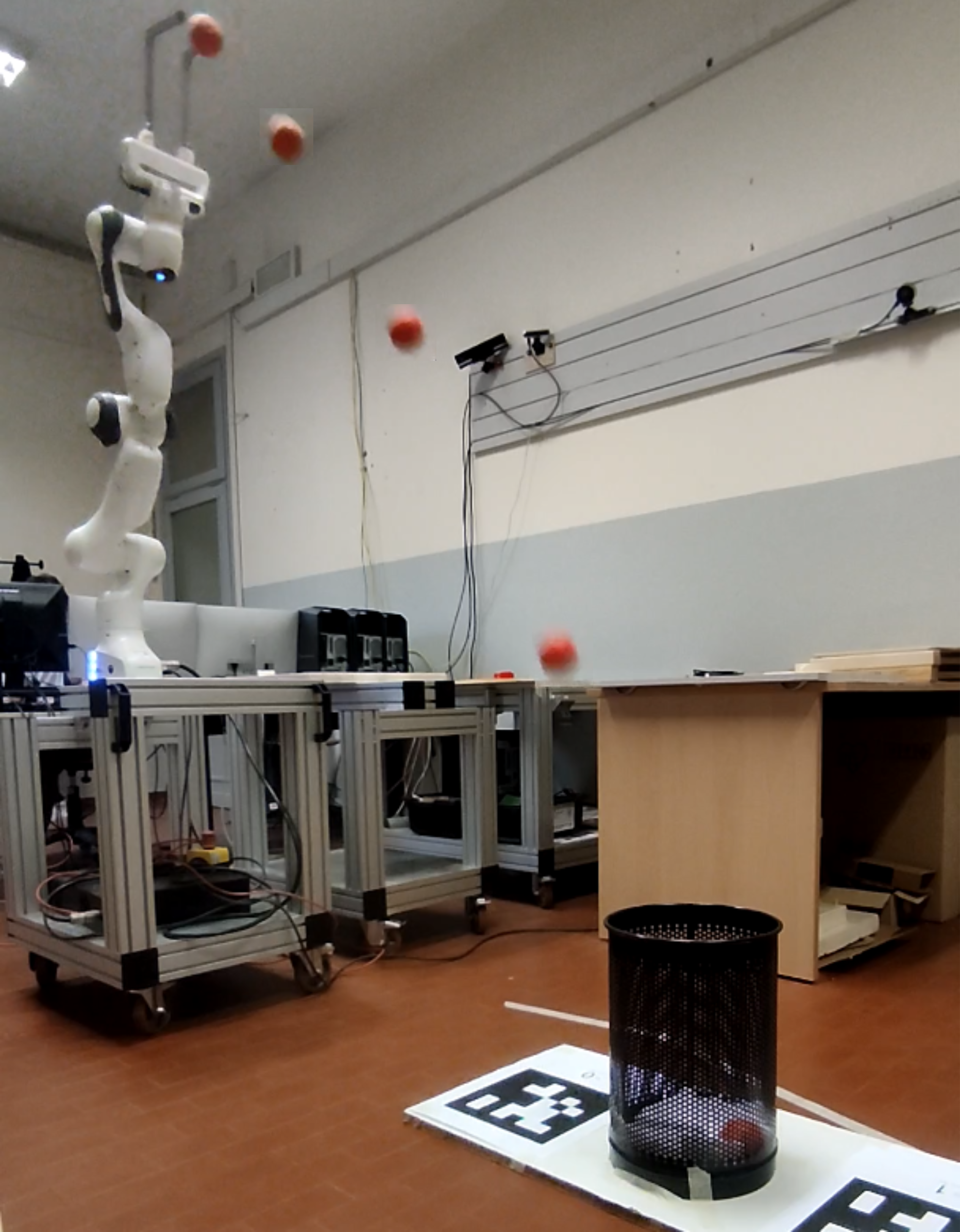}}
    \caption{Panda Robot executing the throwing task with target bin.}
    \label{fig:robot_pic}
\end{figure}

The paper is organized as follows: \Cref{sec:related_work} discusses related work and covers the theoretical background on Gaussian Process Regression (GPR). \Cref{sec:object_throwing_sys} introduces a general robotic object-throwing system and the object-throwing task. \Cref{sec:mc_pilco_obj_throwing} presents the proposed MBRL algorithm MC-PILOT, while \Cref{sec:delay_dist_opt} introduces a possible method to estimate the release delay of the system. \Cref{sec:exp} describes the experimental setup and the achieved results. Finally, \Cref{sec:conclusions} concludes the paper.

A preliminary version of MC-PILOT was introduced in \cite{mcpilco_robot_tossing_sim}. This manuscript improves on the previous conference paper on several aspects: (i) this manuscript analyzes and proposes a solution to estimate object release
delays. (ii) The delay model estimation is included in the Reinforcement Learning loop. (iii) This manuscript presents several experimental results on an actual setup, highlighting the importance of correctly handling the presence of delays to obtain an effective Pick-and-Throw algorithm.

\section{Related Works and Background}
\label{sec:related_work}
This section briefly reviews relevant contributions related to PnT and object-throwing, we divide works into two classes: (i) analytical solutions and (ii) learning-based solutions. Moreover, we report background notions on GPR.

\subsection{Analytical solutions}
In \cite{urban_waste_throwing}, the authors present an innovative industrial-grade waste sorting system composed of a robot performing a PnT task. The paper focuses on demonstrating the overall advantages of PnT over PnP. The PnT strategy is completely engineered, and targets are in fixed positions just outside the robot workspace. The system is composed of a Delta Robot, equipped with a Deep Learning powered vision system, and a conveyor belt where waste material is fed to the robot. The Manipulator has to pick each thrash object from the conveyor, recognize the class of the material, and throw the picked object in the bin assigned to the material. The PnT procedure proves to be faster than a classical PnP, allowing the system to process significantly more waste in the same amount of time.
\cite{taylor2019optimal_motion_planning} introduces a framework for optimizing both the shape and motion of a planar rigid end-effector to meet a specified manipulation task. The framework addresses the optimization and real-world implementation of the shape and motion of a dynamic throwing arm. The task is defined by a set of constraints and a fitness cost, ensuring that the solution aligns with the dynamics of frictional hard contact while fulfilling the task requirements.
In \cite{stochastic_motion_planning_obj_throwing} the authors present an offline motion planning for a robotic arm to execute a throwing task, aiming to deliver an object to a specified target. The planning algorithm seeks a feasible throwing trajectory that adheres to kinodynamic constraints.
\cite{ball_throwing_visual_feedbacK_gripper} is a ball-thrower robotic system, a visual system estimates the target's position, while offline motion planning implements the throwing motion.

\subsection{Learning-based solutions}
Among learning-based solutions, we can make a further distinction based on how objects are released from the manipulator in mid-motion.
Handling the release of objects during arm motion is a complex task for a robotic system since it requires precise synchronization between the arm and the gripper. In several real-life scenarios, the arm and the gripper are not synchronized in real-time, thus making the object tossing particularly challenging. Even small errors can substantially change the final outcome if the arm velocity is high, such as the one required to throw objects. 
For this reason, previous works on robotic skill learning tackle the problem of object-throwing employing specialized cup-like end effectors or placing the objects over the end-effector unconstrained. In this way, the release is obtained by a deceleration of the arm, that effectively detaches the object from the end-effector. This approach simplifies the engineering behind a real-world setup and allows one to focus on the skill-learning or motion-planning task. We first discuss works that rely on cup-like end-effectors, and finally, we present the solutions using a gripper.

\subsubsection{Object-throwing with cup-like end-effector}

In \cite{RL_policy_training_ball_throwing}, the authors use MFRL to train a tossing policy to map raw visual observations to a sequence of velocity commands. The agent receives a reward when the bullet hits or lands close to a target object.
In \cite{hierarchical_RL_ball_throwing} the authors train a simulated 3 Degrees of Freedom (DoF) robot how to throw a ball to a goal position on a hilly landscape, the agent learns a hierarchical policy that decides where the robot is placed in the space between two possibilities, on top of the joint velocity.
In \cite{gutzeit2018besman_ball_throwing}, ball throwing is a benchmark for a skill learning framework, which uses human demonstration to build basic actions and MFRL to refine skill learning.
In \cite{kober2012reinforcement_learning_primitives} MFRL is used to tune motor primitives for a robotic arm that is used to throw a ball at a fixed target.
To the best of our knowledge, \cite{robot_skill_learning_deep_autoencoder_ball_throwing} is the most complete work of this category. The object-throwing task is learned in a latent space of the selected skill representation, allowing generalization to arbitrary target positions. Although results are noteworthy, the real-world experiments only show results for 2 target positions.

Despite these remarkable results, using a cup instead of a gripper poses considerable limitations to the application of PnT in an actual industrial environment, since we can not rely on standard strategies for grasping, which should instead be addressed through ad-hoc hardware or software solutions.

\subsubsection{Object-throwing with gripper grasping}
The task of tossing objects with a robotic manipulator, using a prehensile tool, like a gripper or a suction cup, is a sub-task of PnT. 
Tossingbot \cite{tossingbot} is a MFRL algorithm that exploits an end-to-end self-supervised Deep learning architecture to learn the whole PnT task by trial and error. 
This algorithm uses two dynamics movement primitives 
\cite{paraschos2013probabilistic_primitives_robotics_jan_peters} to parameterize the grasping and throwing skills. 
Two neural networks are trained on experience, to decide grasping and throwing parameters, namely the orientation of the end-effector in the grasping operation, and the release velocity of the throwing primitive. In this way, the release direction is fixed, and the learning-based procedure optimizes only the release velocity. The training is done by maximizing the pick-and-throw success rate based on outcomes collected on the real system, without relying on a model of the system dynamics. 
On one hand, this choice simplifies the algorithm setup, on the other hand, it could limit data efficiency. This means that the algorithm might require a considerable number of trials to effectively learn the task. For instance, empirical results in \cite{tossingbot} demonstrate that even for the task of picking up and throwing a ball, Tossingbot necessitates hundreds or even thousands of trials to reach convergence. Experiments carried out considered a relatively small target variability, namely, targets are a grid positioned in front of the robot. 
In \cite{pick_and_throw_sorting_deep_rl}, the authors introduce a PnT implementation, based on MFRL. A Deep Q-Network (DQN) \cite{DQN} approach is used to evaluate objects' picking feasibility from visual data. The  MFRL algorithm Deep Deterministic Policy Gradient (DDPG) \cite{DDPG} is used to derive a throwing policy. Also in this work, the release direction is fixed, and the throwing policy determines only the release velocity. The method is applied both in simulation and on an actual setup. As in \cite{urban_waste_throwing}, results show potential improvements as concerns time-efficiency, while the operational workspace is marginally enlarged, namely, targets are just outside the robot workspace. As in \cite{tossingbot}, the multiple target areas are a fixed grid.
In \cite{softToss_gripper}, the authors present an object-throwing system implemented with a soft robot \cite{della2020soft_robots}. A MFRL approach based on Proximal Policy Optimization (PPO) \cite{ppo} learns the actuation pattern of the arm, for the tossing task. Also in this work targets are multiple but in fixed locations. \cite{learning_ball_throwing_decision_transformers_gripper} implements an object-throwing system using an industrial manipulator, a MFRL approach based on Decision Transformers (DT) \cite{DT} is used to learn joint trajectories as well as gripper opening time, for the task of throwing a ball toward a single fixed target.

\subsection{Background: Gaussian Process Regression}
\label{sec:gp}


GPR is a machine learning tool used for regression problems, in this work GPR is used to derive a data-driven model of object dynamics.
Let $\D = \{\X, \y\}$ be an input-output dataset, where 
$\X = [{\x}_1^T, \dots, {\x}_n^T]^T$, and $\y=[y_1, \dots y_n]^T$ are, respectively, measurements of the inputs and relative outputs of a system at steps $t = 1 \dots n$.
GPR assumes the following probabilistic model,
\begin{equation}
    \y = f({\X}) + \e = \begin{bmatrix}
        f({\x}_1) \\ \vdots \\ f({\x}_n)
    \end{bmatrix} + \begin{bmatrix}
        {e}_1 \\ \vdots \\ {e}_n
    \end{bmatrix},
\end{equation}
where vector $\e$ accounts for noise, defined a priori as zero mean independent Gaussian noise with variance $\sigma^2$. $f$ is modeled a prior as a zero mean Gaussian Process (GP), with covariance defined by a kernel function $\kappa({\x}_{i},{\x}_{j})$. Namely,  $f({\X}) \sim N(0, K_{{\X} {\X}})$, where the element of $K_{{\X} {\X}}$ at row $i$ and column $j$ is $E[y_{i}, y_{j}] = \kappa({\x}_{i},{\x}_{j})$. 
One typical choice of kernel function is the squared-exponential kernel: 
\begin{equation}\label{eq:se-kernel}
    \kappa({\x}_{i}, {\x}_{j}):= \lambda^2 \e^{-\norm{{\x}_{i}- {\x}_{j}}^2_{\Lambda^{-1}}}
\end{equation}
where $\lambda$ and $\Lambda$ are hyperparameters that can be trained by maximizing the marginal likelihood (ML) of the training samples \cite{rasmussen2003gps_for_ml}.
$\lambda$ is a scaling factor that determines the a priori standard deviation of the target function from its mean. Instead, $\Lambda$ is a matrix that defines the metric used to compute the weighted distance between samples. In this work, we assume $\Lambda$  is a diagonal matrix, with the diagonal elements named lengthscales.

As explained in \cite{rasmussen2003gps_for_ml}, the posterior distribution of $f(x_t)$ given $\mathcal{D}$ is Gaussian distributed, with mean $\E[\hat{f}(x_t)]$ and variance $var[\hat{f}(x_t)]$ expressed by the following closed-form expressions:
\begin{subequations}
    \begin{align}
        &\E[\hat{f}(x_t)] = K_{{\x}_t {\X}} \Gamma_i^{-1} \y    \label{eq:gp_regression_formulas-mean}\\
        &var[\hat{f}(x_t)] = \kappa ({\x}_t, {\x}_t) - K_{{\x}_t {\X}} \Gamma^{-1} K_{{\X} {\x}_t}\label{eq:gp_regression_formulas-var} \\
        &\Gamma = K_{{\X} {\X}} + \sigma^2 I \nonumber
    \end{align}
    \label{eq:gp_regression_formulas}
\end{subequations}
with $ K_{{\x}_t {\X}} = [\kappa(\x_t, \x_1), \dots, \kappa(\x_t, \x_n)]$, and $ K_{{\X} {\x}_t} = K_{{\x}_t {\X}}^T$.
Given the Gaussian distribution, $\E[\hat{f}(x_t)]$ is also the maximum a posteriori estimate of $f(x_t)$. Instead, $var[\hat{f}(x_t)]$ provides a measure of the uncertainty, due to the limited availability of training data or intrinsic stochasticity of the system. 
For the use of GPs in system identification and control we refer the interested reader to
\cite{kernel_methods_and_gp_control_systems_magazine}.

\section{Robotic Object throwing}
\label{sec:object_throwing_sys}
This section provides a general specification of a robotic object-throwing system, including a geometrical description of the object-throwing task, requirements of the robot motion, and definition of a baseline policy for the system.

\subsection{Problem definition}
The object-throwing task can be described as follows. Consider an object in a known position, inside the reachable workspace of a robotic manipulator. The manipulator must grasp the object and toss it inside a target bin, whose location is variable, but known and selected within a pre-specified domain. As in most of the previous works, we assume the manipulator releases the object from a pre-computed configuration that depends on the position of the bin. 

\begin{figure}[h]
	\centering
	\subfloat{\includegraphics[width=\columnwidth]{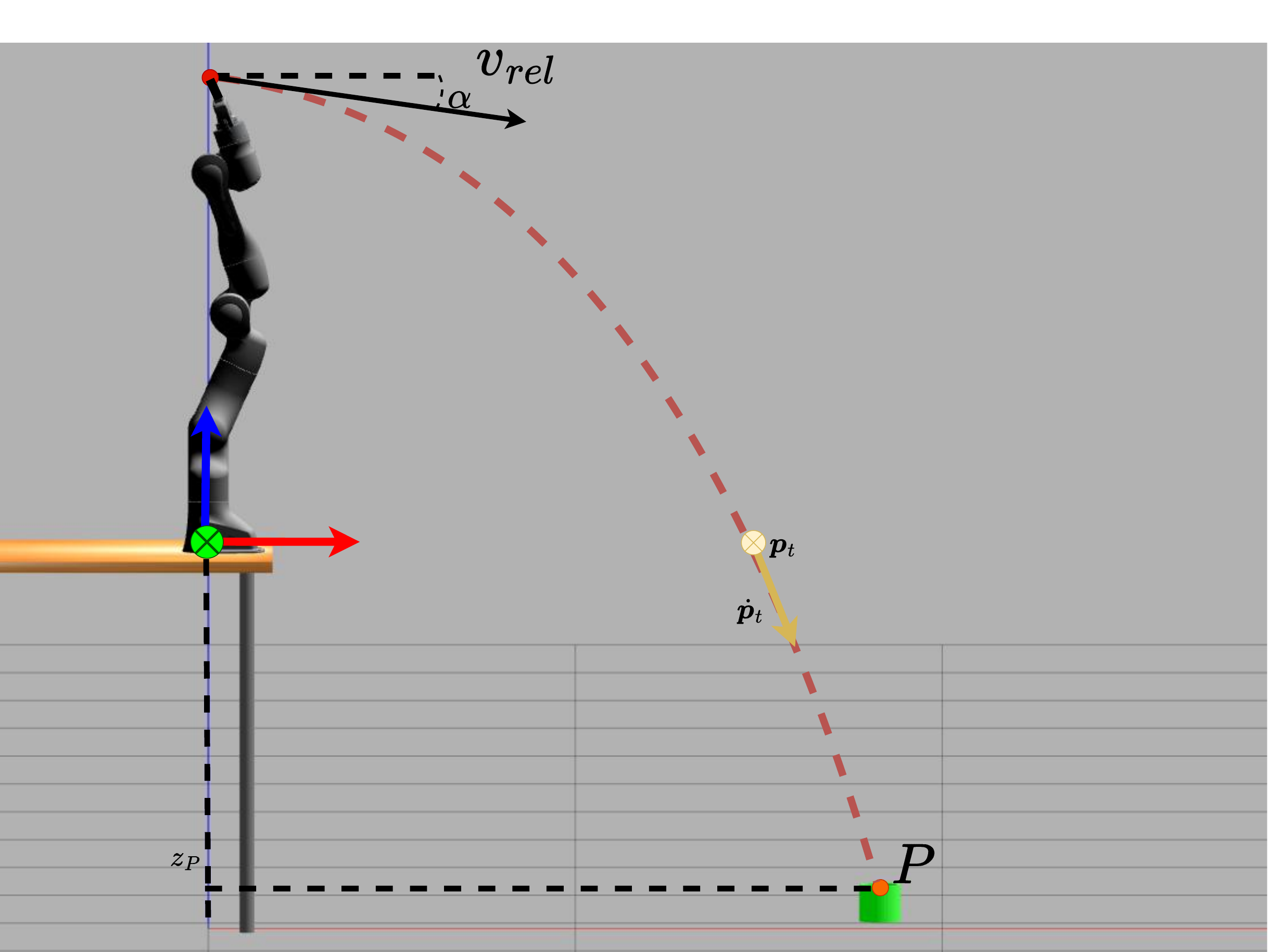}}
	
	\subfloat{\includegraphics[width=\columnwidth]{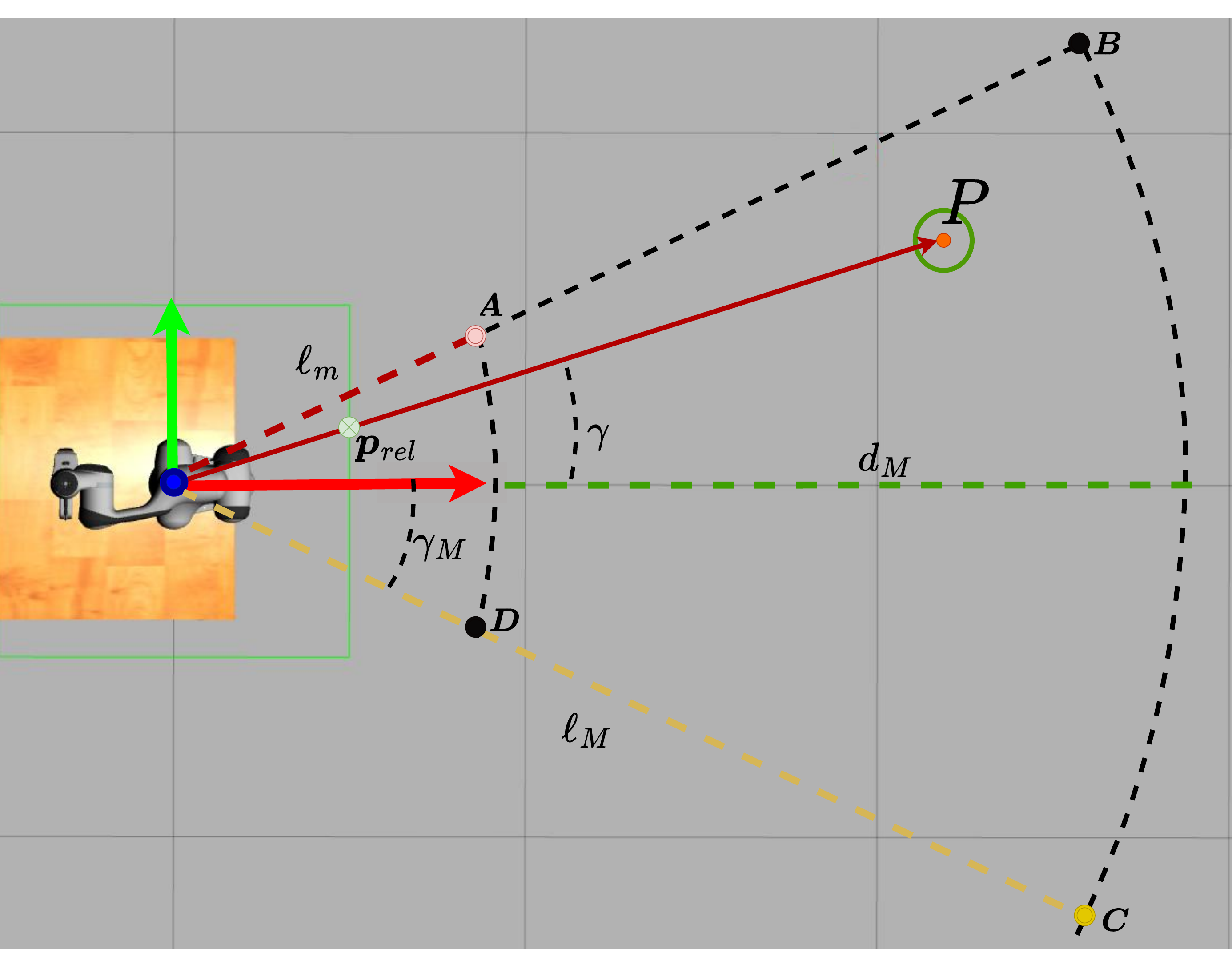}}
	\caption{Screenshots from the simulation, robot in release configuration. 
		The bin is placed in an example position. The target position $\Pb$ is the top opening of the cylinder. 
		The RBG axis triplet is the robot's reference frame.}
	\label{fig:panda_sim}
\end{figure}

\Cref{fig:panda_sim} describes the ideal geometry of the task, while a description of the actual setup is reported in \Cref{sec:exp}. Let us assume that the global reference frame coincides with the robot's base frame, denoted as $\boldsymbol{O}-xyz$. The bin's location is inside the area $\boldsymbol{ABCD}$. As described in \Cref{fig:panda_sim} (bottom plot), this area belongs to a plane of height $z_P$ and it is comprised between two circular arcs centered in $\boldsymbol{O}$ of radius $\ell_m$ and $\ell_M$, respectively, and having minimum and maximum angular amplitudes $-\gamma_M$ and $\gamma_M$. Hence the domain of the target $\Pb$, namely, the bin's center, is 
\begin{equation}
    \D_P = \left\{
    \begin{bmatrix}
        x_P \\
        y_P \\
        z_P
    \end{bmatrix}
    = 
    \begin{bmatrix}
        \ell \cos{(\gamma)} \\
        \ell \sin{(\gamma)}\\
        z_P
    \end{bmatrix}
    \text{ with   }
    \begin{aligned}
        &\ell \in [\ell_m, \ell_M],\\
        &\gamma \in [-\gamma_M, \gamma_M]
    \end{aligned}
    \right\} 
    \label{eq:domain}
\end{equation}

Regarding the release position and velocity, we consider the approach followed by several previous works, see, for instance, \cite{tossingbot}. In this strategy, the position and the release direction are completely determined by the target position $\Pb$, and the robot optimizes only the release velocity module. Specifically, the robot releases the object from a point with fixed height $z_{rel}$, distant $\ell_r$ from the center, and with a velocity vector that is aligned with $\gamma$. Then, given the target position $\Pb$, and its representation in polar coordinates $[\ell, \gamma]$, the release position $\p_{rel}$ is 
\begin{equation}
\p_{rel}= 
\begin{bmatrix}
    \ell_r \cos{(\gamma)} & \ell_r \sin{(\gamma)} & z_{rel}
\end{bmatrix}^T.    
\label{eq:release_point}
\end{equation}
Instead, the direction of the tossing velocity $\velbf_{rel}$ is expressed by the unity vector $\velbf_\gamma^\alpha$, defined as
\begin{equation}
    \velbf_\gamma^\alpha = 
    \begin{bmatrix}
        \cos{\alpha} \cos{\gamma} &
        \cos{\alpha} \sin{\gamma} &
        \sin{\alpha} 
    \end{bmatrix}^T
    \label{eq:vel_dir},
\end{equation}
where $\alpha$ is the vertical angle between $\velbf_{rel}$ and the ground, and it is assumed constant and given; see the top plot in Figure \ref{fig:panda_sim}.

Then, the release velocity is 
\begin{equation}
    \velbf_{rel} =  \vel \velbf_\gamma^\alpha,
    \label{eq:vel_rel}
\end{equation}
where $\vel$ is the velocity magnitude selected by the tossing algorithm. 

\subsection{Motion planning}
\label{subsec:motion_plan}
This section details the requirements of a motion planning algorithm for the object-throwing system, given the geometric description reported in \cref{fig:panda_sim}.
Suppose that the release position $\p_{rel}$, given target $\Pb$, is reachable by the robot for a certain configuration $\q_{rel}$, i.e.
\begin{equation}
    \p_{rel} = f_{kin} (\q_{rel}),
\end{equation}
where $f_{kin}$ is the forward kinematics function. Let $\q_t$ and $\dq_t$ be the position and velocity of the manipulator's joints at time $t$. To throw the object from $\p_{rel}$ with Cartesian velocity $\velbf_{rel}$, as defined in \cref{eq:vel_rel}, we have to design a joint trajectory $\q_t^\vel$ and $\dq_t^\vel$, with $t\in [0, T]$, such that, at a given time $t_r$, hereafter referred as release time, the following relations hold:
\begin{equation}
    \begin{split}
        \q_{t_r}^\vel &= \q_{rel} \\
        \velbf_{rel} &= J_a(\q_{rel})\dq_{t_r}^\vel = \vel \velbf_\gamma^\alpha
    \end{split}
\end{equation}
where $J_a$ is the robot's analytical Jacobian defined as
\begin{equation}
    J_a (q) = \frac{\partial f_{kin}(q)}{\partial q} = \begin{pmatrix}
        \frac{\partial f_{kin}(q)}{\partial q_1} & \dots & \frac{\partial f_{kin}(q)}{\partial q_{dof}}
    \end{pmatrix}.
\end{equation}

In this way, the robot accelerates from an initial configuration and reaches the release configuration $\q_{rel}$ with cartesian velocity $\velbf_{rel}$, at time $t_r$. Finally, the robot has to decelerate to stop the motion. The kinodynamic constraints of the manipulator determine the maximum tossing velocity. If the desired velocity is inside the robot limits, we can assume that the manipulator follows the desired trajectory with negligible tolerance.

When the manipulator reaches the configuration $\q_{rel}$, the robot gripper must be open to release the object such that the object dynamics switch to free motion. Unfortunately, most grippers are not included in the robot real-time control loop, this opening command is typically executed with a certain delay, that may or may not be estimated. This means that, in general, the opening command should be forwarded at time $t_{r_{cmd}}$, before the nominal release time $t_r$, to compensate for the delay; namely, $t_{r_{cmd}}<t_r$. Nonetheless, these kinds of delays are difficult to estimate and are typically not entirely deterministic, meaning that it's likely that a complete compensation of the delay is not possible. Furthermore, the actual object release depends also on the geometrical and physical object characteristics and their interaction with the gripper. As a result, even assuming a perfect synchronization between the manipulator and the gripper, it is extremely hard to obtain a perfect object release. 

These nonideal conditions can be synthesized by assuming that the object detaches from the gripper at an unknown time $\tilde{t}_r$, defined as 
\begin{equation}
 \tilde{t}_r = t_{r_{cmd}} + t_d 
 \label{eq:actual-release-time}
 \end{equation}

where $t_d$ is the unknown delay. Since we assumed that the robot follows the reference trajectory precisely, the actual release happens when the robot is in configuration $\q_{\tilde{t}_r}^\vel$, $\dq_{\tilde{t}_r}^\vel$, which in cartesian space translates to 
\begin{equation}
 \begin{split}
    &\tilde{\p}_{rel} = f_{kin} (\q_{\tilde{t}_r}^\vel) \\
    &\tilde{\velbf}_{rel} =  J_a(\q_{\tilde{t}_r}^\vel)\dq_{\tilde{t}_r}^\vel= \tilde{\vel} \tilde{\velbf}_{\gamma}^{\alpha}
 \end{split}
\end{equation}
i.e. the release point is different from the desired one, and, consequently, also the cartesian velocity is incorrect both in direction and magnitude.

\subsection{Baseline Policy}
\label{subsec:baseline_policy}
Supposing that the target's location $\Pb$ is either known or measured precisely enough, then a policy for this system is a function $\pi: \R^3 \rightarrow \R$ that takes as input $\Pb$ and returns the desired release velocity $\vel$.
Under ideal conditions, namely, modeling the objects as a point mass and neglecting complex behaviors such as friction and delays, $\pi $ could be obtained in closed form exploiting standard ballistic equations, namely 
\begin{equation}
    \pi(P) = \sqrt{\frac{g \cdot d^2}{2\cos^2{(\alpha)}(d\tan{(\alpha)} - z_P +z_{rel})}}
    \label{eq:balistic-no-frict}
\end{equation}
where 
$d = \sqrt{(x_P - \ell_r \cos{(\gamma)})^2 + (y_p - \ell_r \sin{(\gamma))^2}}$
, $g$ is the acceleration due to gravity, and $\gamma = \text{ATAN2}(y_P, x_p)$.

The errors due to the above approximations make the solution based on \cref{eq:balistic-no-frict} too imprecise. 
Moreover, further errors come from issues related to the implementation of the robot throwing: the robot controller follows the reference trajectory with a certain precision, but at the same time, uncompensated release delays cause relevant mistakes, as described in \Cref{subsec:motion_plan}. 

\begin{figure}[t]
    \centering
    \subfloat{\includegraphics[width=0.8\columnwidth]{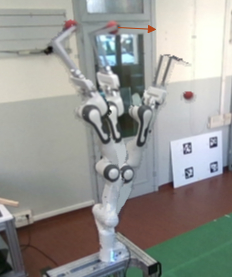}}
    \caption{The robot performs a throwing motion. 
    }
    \label{fig:throwing_motion}
\end{figure}

\section{MC-PILOT}
\label{sec:mc_pilco_obj_throwing}
This section describes the MBRL algorithm developed to train the robot for the object-throwing task. The algorithm is called MC-PILOT and is derived from MC-PILCO \cite{mcpilco_tro}. The pseudocode of MC-PILOT is reported in \Cref{alg:mc_pilot}. MC-PILOT is a Model-Based policy gradient algorithm, in which GPR is used to estimate system dynamics, and long-term state distributions are approximated via Monte Carlo simulation.
\begin{remark}
From this point forward, we use the variable $\tp$ to identify the simulation time of MC-PILOT, instead of the robot time $t$.
\end{remark}
MC-PILOT retains the architecture of MC-PILCO, but important adaptations were devised to solve the object-throwing task. Two main differences between the PnT problem and the tasks typically considered in MC-PILCO entail substantial algorithmic modifications. 

(i) First, the policy acts on the system only at $\tp=0$ by selecting the desired velocity, while for $\tp>0$ the object is in free motion. This means that the policy's control action affects the initial state distribution, and then the state evolution depends only on the previous state. Then, we can describe the system's evolution as,
\begin{equation}
    \begin{cases}
        \x_0  \sim p_0(u_0), \\
        \x_{\tp+1} = f (\x_{\tp}) + \w_{\tp}.
    \end{cases}
    \label{eq:system_equation_tossing}
\end{equation}
where the system input $u_0$ is the release velocity, $\x_{\tp} \in \R^{d_x}$ is the state of the system at time $\tp$, $f: \R^{d_x} \rightarrow \R^{d_x}$ is the discrete-time unknown transition function, while $\w_{\tp}$ is an independent white noise describing uncertainty influencing the system evolution. The state is defined as $\x_{\tp}= [\p_{\tp}^T, \dpb_{\tp}^T]^T$, with $\p_{\tp},\dpb_{\tp} \in \R^3$ pointing out, respectively, the cartesian position and velocity of the object's center of mass at time $\tp$.


(ii) Second, we want to learn simultaneously to throw objects in any target configuration in $\D_P$, namely, the target is not fixed. To allow for simultaneous learning of the PnT task to any target in $\D_P$ we define an augmented state $\tilde{\x}_{\tp}$ as
\begin{equation}
    \tilde{\x}_{\tp}  = 
    \begin{bmatrix}
        \x_{\tp}^T&
        \Pb^T
    \end{bmatrix}^T
    \label{eq:extended_state}
\end{equation}
where $\Pb \in \D_P$ collects the Cartesian coordinates of the target. With this formulation, it is possible to define the policy and the cost as functions of the extended state $\tilde{\x}_{\tp}$. The policy is a function $\pi_{\thetab}: \D_P \rightarrow \R$, which depends on the parameters $\thetab$ and selects the input to be applied at the initial state, based on $\Pb$. The cost, instead, is a function that penalizes the distance between the target and the object's position. The cost function we used to encode the task is the saturated distance from the target state:
\begin{align}
    \begin{split}
        c_{st}(\tilde{\x}_{\tp}) = 1 - e^{- \norm{\p_{\tp} - \Pb}^2_{\Sigma_c}}   \hspace{0.5cm}
        \Sigma_c = \text{diag}\left(\frac{1}{\ell_c}, \frac{1}{\ell_c}, 0\right)
    \end{split}
    \label{eq:saturated_dist_target}
\end{align}
Furthermore, the objective function minimized to update the policy is:
\begin{equation}
    J(\thetab) = \E_{p_0, \pi_{\thetab}} [c_{st}(\tilde{\x}_T)]
    \label{eq:cumulative_cost_tossingbot}
\end{equation}
that is the expectation of $c_{st}(\tilde{\x}_{T})$, which encodes the object-throwing task. Indeed, in the PnT task, we are interested only in the landing position, i.e., the final state of the system, and not in the trajectory performed by the object. The objective is to find $\thetab^*$ that minimizes $J(\thetab)$.

MC-PILOT performs a succession of attempts to solve the desired task, also called trials. The trials consist of three main phases: (i) model learning, (ii) policy update, and (iii) policy execution.
The model learning step uses collected experience to build or update the GP model. Training data is extracted from free-motion object trajectories measured through a motion capture system, which also measures the landing points. The policy update step formulates an optimization problem to minimize the objective $J(\thetab)$ in \cref{eq:cumulative_cost_tossingbot} w.r.t. the parameters of the policy $\thetab$. 
At the end of the trial, the current optimized policy is tested on the system and the interaction data is stored to update the model for the following trial. At the first trial, the algorithm is initialized with data collected using an exploration policy, for instance, a random policy, or the baseline policy from \Cref{eq:balistic-no-frict}. \\
In the following, we detail the three main phases of MC-PILOT.

\begin{algorithm}[t]
	\caption{MC-PILOT}
	\label{alg:mc_pilot}
	\begin{algorithmic}[5]
		\STATE // Initial exploration
		\FOR{$j = 1, \dots, N_{exp}$}
		\STATE Sample $\Pb \sim \D_P$
		\STATE Throw towards $\Pb$ with the baseline policy
		\STATE Collect the object's trajectory and perform data augmentation 
		\STATE Add new $N_a+1$ trajectories to the model's dataset       
		\ENDFOR
		\STATE Initialize parameters (from \cref{table_hyperparam})
		\WHILE{Task is not learned}
		\STATE // Model Learning (\Cref{subsec:mc_pilot_model})
		\STATE Train the GP model 
		\STATE // Delay estimation (\Cref{sec:delay_dist_opt})
		\STATE Estimate $(a,b)$ 
		
		\STATE // Policy Update (\Cref{subsec:policy_update})
		\FOR{$j = 1 \dots N_{opt}$}
		\FOR{$m = 1 \dots M$}
		\STATE $\Pb^{(m)} \sim \D_P$ and $t^{(m)}_d \sim U(a,a+b)$
		\STATE Compute $\x_0^{(m)}(\pi_{\thetab}(\Pb^{(m)}))$
		\ENDFOR
		
		\FOR{$\tp = 0 \dots T-1$}
		\STATE $\x_{\tp+1}^{(m)} \sim  \mathcal{N} (\mub_{\tp+1}^{(m)}, \Sigma_{\tp+1}^{(m)})^T$, for $m = 1 \dots M$
		\ENDFOR
		
		\STATE Compute $\hat{J}(\boldsymbol{\thetab})$ from \cref{eq:cost_estimate_monte_carlo_tossing}
		\STATE Update $\thetab$ with $ \nabla_{\thetab} \hat{J}(\boldsymbol{\thetab})$
		\ENDFOR
		
		\STATE // Policy Execution
		\FOR{$j = 1, \dots, N_{test}$}
		\STATE Throw towards $\Pb \sim \D_P$ with $\pi_{\thetab}$
		\STATE Add new $N_a+1$ trajectories to the dataset.
		\ENDFOR

		\ENDWHILE
	\end{algorithmic}
\end{algorithm}

\subsection{Model Learning}
\label{subsec:mc_pilot_model}
In the model learning phase, GPR is used to estimate a model of system dynamics, i.e. a stochastic estimate of the transition function in \Cref{eq:system_equation_tossing}. We employ the speed-integration model from \cite{mcpilco_tro}. 
Assuming constant accelerations between successive time steps, one can compute the state evolution as 
\begin{equation}
    \label{eq:speed-int}
    \begin{split}
        &\dot{\p}_{\tp+1} = \dot{\p}_{\tp} + \Delta_{\tp} \\
        &\p_{\tp + 1} = \p_t + T_s \dot{\p}_{\tp} + \frac{T_s}{2} \Delta_{\tp}
    \end{split}
\end{equation}
where $T_s$ is the sampling time and  $\Delta_{\tp} = \dot{p}_{\tp+1} - \dot{p}_{\tp}$ is the vector that collects the velocities changes.
MC-PILOT estimates the unknown functions $\Delta_{\tp}$, which model the change in velocity, from collected data $\mathcal{D}$ employing GPR.  Each component of  $\Delta_{\tp}$ is modeled as an independent GP with the same input $\tilde{\x}_{\tp}$. The GPs are equipped with a squared-exponential kernel of the kind in \Cref{eq:se-kernel}. Each collected trajectory $\X$ is included in the training dataset using a simple data augmentation procedure: an arbitrary number of $N_a \in \Nat$ trajectories is derived from $\X$ by considering a random rotation around the vertical axis. In this way, we exploit symmetries and improve data efficiency.
Then, through GPR, MC-PILOT obtains in closed form a model of the one-step ahead transition function in \eqref{eq:system_equation_tossing}.
Based on \eqref{eq:speed-int} and \eqref{eq:gp_regression_formulas}, and assuming that the system is in state $\x_{\tp}$, the one-step-ahead dynamics is Gaussian distributed, namely, 
\begin{equation}
    p(\x_{\tp+1} | \x_{\tp}, \D) \sim \mathcal{N} (\mub_{t+1}, \Sigma_{t+1}) \label{eq:one-step-posteior}
\end{equation}
with mean $\mub_{\tp+1}$ and covariance $\Sigma_{\tp+1}$.
Specifically, since the GP models are independent of each other, given the input $\tilde{\x}_{\tp}$, also the posterior distributions are independent Gaussians, with mean and variance computed in closed form from \eqref{eq:gp_regression_formulas}. Then, the posterior distribution of $\Delta_{\tp}$ is Gaussian, with mean $E[\Delta_{\tp}]$ and covariance matrix $\Sigma[\Delta_{\tp}]$, where $E[\Delta_{\tp}]$ collects the GPs' posterior mean computed as in \eqref{eq:gp_regression_formulas-mean}, while $\Sigma[\Delta_{\tp}]$ is a diagonal matrix with the posterior variances from \eqref{eq:gp_regression_formulas-var}. Finally, the posterior mean and covariance in \eqref{eq:one-step-posteior} are obtained from \eqref{eq:speed-int}, exploiting properties of Gaussian distribution and linear transformation. Since $\Delta_{\tp} \sim \mathcal{N}(E[\Delta_{\tp}], \Sigma[\Delta_{\tp}])$ and \eqref{eq:speed-int} is linear, we obtain,
\begin{equation}
\begin{split}
    &\mub_{\tp+1} = \begin{bmatrix}
    p_{\tp}\\
    \dot{p}_{\tp}
\end{bmatrix}
+ 
\begin{bmatrix}
    T_s \dot{p}_{\tp} + \frac{T_s}{2} \E[\Delta_{\tp}] \\
     \E[\Delta_{\tp}]
\end{bmatrix}    \\
    &\Sigma_{\tp+1} = \begin{bmatrix} 
    \frac{T_s^2}{4} \Sigma[\Delta_{\tp}] &\frac{T_s}{2} \Sigma[\Delta_{\tp}]  \\ \frac{T_s}{2} \Sigma[\Delta_{\tp}] &  \Sigma[\Delta_{\tp}] \end{bmatrix}.
\end{split}
\label{eq:posterior_dist_speed_int}
\end{equation}

\subsection{Policy update}
\label{subsec:policy_update}
In the policy update phase, the policy is trained to minimize \cref{eq:cumulative_cost_tossingbot} with the expectation computed w.r.t. \cref{eq:one-step-posteior}, starting from the initial distribution $p_0(u_0)$. This computation is not possible in closed form, therefore as done in MC-PILCO, MC-PILOT performs an approximation based on Monte Carlo sampling \cite{caflisch1998monte_carlo_sampling_ref}. Remarkably, the use of Monte Carlo sampling allows to employ any kind of initial distribution, which is fundamental for this work.
First, we define the policy structure and its parameters, second, we describe the Monte Carlo optimization procedure.

\subsubsection{Policy definition}
The policy function depends only on the augmented part of the state, namely, $\Pb$. As a policy function, we adapted the general purpose squashed radial basis function network presented in \cite{mcpilco_tro}, but any parametric function can be used. Then, the desired release velocity is obtained as, 
\begin{align}
    \begin{split}
         \pi_{\thetab}(\Pb) &= \frac{u_{M}}{2} \left(  \tanh{ \left(  \sum_{i=1}^{N_b} \frac{w_i}{u_{M}} e^{-\| \ab_i - \Pb \|^2_{\Sigma_\pi}}  \right)} +1 \right) \\
    \end{split}
    \label{eq:tossing_policy}
\end{align}
where $u_M$ is the maximum release velocity, while $\thetab = \{{\bf w}, A, \Sigma_\pi\}$ is the set of policy parameters to be optimized; ${\bf w} = [w_1, \dots, w_{N_b}]$ and $A= \{\ab_1, \dots, \ab_{N_b}\}$ are, respectively, weights and centers of the $N_b$ Gaussians basis functions, whose shapes are determined by $\Sigma_{\pi}$. Notice that the squashing function $\tanh()$ keeps the output between zero and the maximum release velocity $u_M$, thus accounting automatically for input constraints. In the experiments, the parameters are initialized as follows. $u_{M}$ is set according to the manipulator's Cartesian limits.
The matrix $\Sigma_{\pi}$ is a constant diagonal matrix.
The basis weights are sampled uniformly in $[-u_M, u_M]$, the centers are sampled uniformly in a superset of $\D_P$ defined as 
\begin{equation}
    \Bar{\D}_P = \left\{
    \begin{bmatrix}
        x_P \\
        y_P \\
        z_P
    \end{bmatrix}
    \text{ with }
    \begin{aligned}
        & x_P \in [0, \ell_M],\\
        & y_P \in [-\ell_M\sin{\gamma_M}, \ell_M\sin{\gamma_M}]
    \end{aligned}
    \right\} 
\end{equation}
This is done to ensure that the centers cover the input region of the policy function as much as possible.

\subsubsection{Monte Carlo Optimization procedure}
The policy optimization procedure implements a gradient-based optimization to update the policy parameters $\theta$. At each step, the MC-PILOT approximates the objective function in \Cref{eq:cumulative_cost_tossingbot} with Monte Carlo sampling \cite{caflisch1998monte_carlo_sampling_ref} and computes the gradient w.r.t. $\theta$. The procedure samples from the initial distribution $p_0(u_0)$ a batch of $M$ particles and simulates their evolution based on \cref{eq:one-step-posteior} and the current policy.
With respect to MC-PILCO, in MC-PILOT the particles' simulation is modified to account for the system dynamics in \eqref{eq:system_equation_tossing} and the complexity of the initial state distribution.
Namely, to draw samples from $p_0(u_0)$, for each of the $M$ particles, first the algorithm uniformly samples a target position $\Pb^{(m)}=[x_P^{(m)}, y_P^{(m)}, z_P^{(m)}]^T$ from $\D_P$, with $m=1 \dots M$. Then, the policy computes the desired Cartesian release velocity $\vel^{(m)} = \pi_\theta(\Pb^{(m)})$ as in \eqref{eq:tossing_policy}. This determines univocally the joints reference trajectory $\q^{(m)}_{t}$ $\dq^{(m)}_{t}$ as described in \Cref{sec:object_throwing_sys}. 

Defining $\x_{\tp}^{(m)}$ as the state of the $m$-th particle at time $\tp$, in the absence of tracking errors and de-synchronizations, and assuming that the object is released at time $t_r$, the release state $\x_0^{(m)}$ would be:
\begin{equation}
    \x_0^{(m)}(\pi_{\thetab}(\Pb^{(m)})) = h(\Pb^{(m)}, \vel^{(m)}, t_r) = \begin{bmatrix}
        f_{kin}(\q^{(m)}_{t_r}) \\
        J_a(\q^{(m)}_{t_r}) \dq^{(m)}_{t_r}
    \end{bmatrix}
    \label{eq:noiseless_particles_init}
\end{equation}
However, due to several factors, discussed in \Cref{sec:object_throwing_sys}, the release time is not deterministic and therefore is affected by uncertainties.
To account for this behavior, we define the actual release time as $t_r = t_{r_{cmd}} + t_d$, with $t_d$ modeled as a uniform distribution. Namely, for each particle, we sample $t^{(m)}_d$ from a uniform distribution $U(a, a+b)$, thus obtaining a sample of the actual release time $\tilde{t}^{(m)}_r = t_{r_{cmd}} + t^{(m)}_d$. Then, the sample of the object's initial state is:
\begin{equation}
    \x_0^{(m)}(\pi_{\thetab}(\Pb^{(m)})) = h(\Pb^{(m)}, \vel^{(m)}, \tilde{t}^{(m)}_r) = \begin{bmatrix}
        f_{kin}(\q^{(m)}_{\tilde{t}_r}) \\
        J_a(\q^{(m)}_{\tilde{t}_r}) \dq^{(m)}_{\tilde{t}_r}
    \end{bmatrix}
\label{eq:particles_init_noise}
\end{equation}
Properly selecting the $t_d$ distribution is crucial to the algorithm's success. In most commercial systems, available measurements are not adequate to directly estimate this distribution in a data-driven fashion since the gripper and the arm are not synchronized. In \Cref{sec:delay_dist_opt}, we propose a method to estimate the pair $(a,b)$, using the GP model and the robot kinematics, thus demonstrating an additional advantage of the Model-based approach.

Then, the particles are simulated for a time horizon of length $T \in \Nat$. The update rule for each time step $t=0, \dots, T-1$ is:
\begin{equation}
    \begin{split}
        \x_0^{(m)} &= \x_0^{(m)}(\pi_{\thetab}(\Pb^{(m)})),  \hspace{0.2cm} \Pb^{(m)} \sim U(\D_P)\\
        \x_{\tp+1}^{(m)} &\sim  \mathcal{N} (\mub_{\tp+1}^{(m)}, \Sigma_{\tp+1}^{(m)}),  \hspace{0.2cm}
        \tilde{\x}_{\tp+1}^{(m)}  = 
            \begin{bmatrix}
                \x_{\tp+1}^{(m)}\\
                \Pb^{(m)}
            \end{bmatrix}^T
        \end{split}
    \label{eq:particles_simulation}
\end{equation}
for $m=1, \dots, M$, where $\mub_{\tp+1}^{(m)}$ and $\Sigma_{\tp+1}^{(m)}$ are computed as described in \cref{eq:posterior_dist_speed_int}, from the input vector $\x_{\tp}^{(m)}$. 
Finally, the evolution of each particle is simulated independently for $T$ steps, following \Cref{eq:particles_simulation}, using the GP dynamics model. 
The sampled targets $\Pb^{(m)}$ are retained throughout the simulated horizon, in the particles' extended states, as defined in \cref{eq:extended_state}.
Since each particle is independent of each other, the computations can be done in parallel. 
The simulation stops when the particle reaches target altitude $z_P^{(m)}$. Namely, if the particle reaches $z_P^{(m)}$ or below at time step $\tp_z$, then for the rest of the samples, the state update equation is changed to $\tilde{x}_{\tp+1}^{(m)} = \tilde{x}_{\tp}^{(m)}$ for $\tp>=\tp_z$. 
The particle's landing position is approximated to the terminal point of its simulated trajectory $\p_T^{(m)}$ in  $\x_T^{(m)}$. 

Then, the expectations in \cref{eq:cumulative_cost_tossingbot} are approximated as the sample mean of the simulated particles' costs, resulting in the following approximation of the objective function:

\begin{equation}
    \begin{split}
        \hat{J}(\boldsymbol{\thetab}) =  \frac{1}{M} \sum_{m=1}^M c_{st} \left( \tilde{\x}_T^{(m)} \right)  ,
    \end{split}
    \label{eq:cost_estimate_monte_carlo_tossing}
\end{equation}
i.e. the mean of the cost of the batch of particles at time $T$, thus considering only the landing position.
The minimization of \cref{eq:cost_estimate_monte_carlo_tossing} w.r.t. $\thetab$ is interpreted as a stochastic gradient descent problem (SGD) \cite{bottou2010large_scale_learning_sgd}, employing the reparameterization trick to differentiate stochastic operations \cite{kingma2013reparametrization_trick}, and solved with an optimizer like Adam \cite{kingma2014adam}.
The reparameterization trick allows differentiation through stochastic operations by sampling from a properly dimensioned zero-mean and unitary variance Gaussian distribution, instead of sampling directly from the original Gaussian distribution. Specifically, each particle is mapped to:
\begin{subequations}
    \begin{align}
        &\x_{t+1}^{(m)} = \mub_{t+1}^{(m)} + L_{t+1}^{(m)} \varepsilon, \\
        &L_{t+1}^{(m)} L_{t+1}^{(m)T} = \Sigma_{t+1}^{(m)},
    \end{align}
    \label{eq:one_step_ahead_repram}
\end{subequations}
where $ \varepsilon \sim \mathcal{N}(0, I_{d \times d})$, and $L_{t+1}^{(m)}$ is the Cholesky decomposition of $\Sigma_{t+1}^{(m)}$. 
As introduced in \cite{mcpilco_tro}, we employ dropout \cite{srivastava2014dropout} of the policy parameters $\thetab$ to improve exploration and increase the ability to escape from local minima during policy optimization. 
The pseudocode of MC-PILOT is reported in \cref{alg:mc_pilot}. 
In the algorithm, we use the variable $N_{exp}$ to indicate the number of exploration throws performed at the beginning of each episode. At the same time, $N_{opt}$ is the number of optimization steps performed in the policy optimization procedure.


\begin{figure*}[h]
    \centering
    \includegraphics[width=\textwidth]{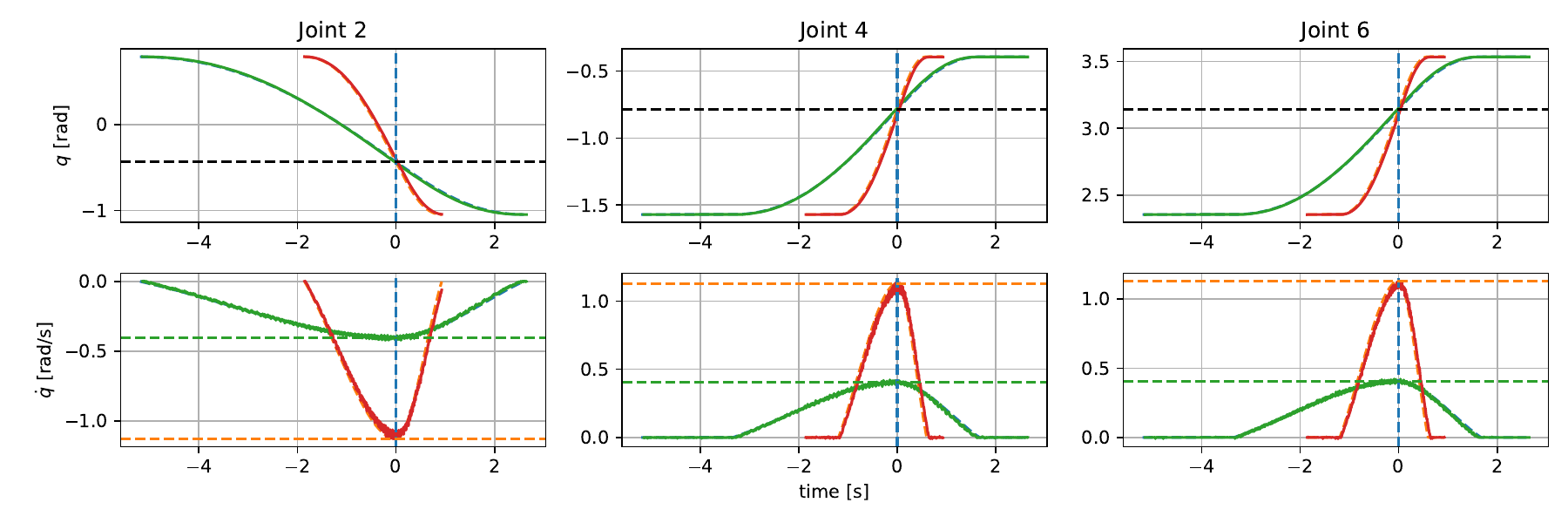}
    \caption{Reference joint trajectory and actual trajectory recorded on the robot, for the tossing motion. The trajectory moves only three joints. Plot shows the trajectories related to the minimum (green) and maximum (red) cartesian velocity.
    The vertical dashed lines show the nominal release time. The horizontal dashed lines are desired joints position and velocity at release.
    }
    \label{fig:reference_trajectories}
\end{figure*}

\section{Delay Distribution Optimization}
\label{sec:delay_dist_opt}
As pointed out in \Cref{subsec:motion_plan}, compensating the opening delay of the gripper $t_d$, eventually considering also stochastic effects, is fundamental to obtain an effective PnT system. In this section, we present a strategy for estimating the parameters of the $t_d$ distribution, assuming that $t_d \sim U(a, b)$. However, the proposed strategy can be easily extended to any parametric distribution.  

This procedure is directly integrated into the MC-PILOT algorithm, it is executed after the model learning step. For each throw performed during exploration or policy execution, we extract from the correspondent free-motion trajectory the actual landing position of $\p^i$, along with the desired release velocity given to the system $\vel_i$ and the target point $\Pb_i$. Therefore, given $N$ experiments, we obtain the following input-output dataset:
\begin{equation}
 D = \begin{bmatrix}
    \left(\left(\vel_1, \Pb_1\right), \p^1\right) \\ 
    \vdots \\
    \left(\left(\vel_{N}, \Pb_{N}\right), \p^{N}\right)
 \end{bmatrix}    
\end{equation}
The proposed strategy combines the dataset $D$ and the GP models to define an optimization problem w.r.t. the parameters of the $t_d$ distribution. For each of the $N$ throws, we perform a Monte Carlo simulation composed of $M_d$ particles. Consider particle $m$ of throw $i$. The particle is initialized as in \cref{eq:particles_init_noise}, with $\vel^{(m)} = \vel_i$, $\Pb^{(m)} = \Pb_i$ and $\tilde{t}^{(m,i)}_r$ computed based on \eqref{eq:actual-release-time} assuming $t_d \sim U(a, a+b)$, namely, $$\tilde{t}^{(m,i)}_r \sim U(t_{r_{cmd}}+a, t_{r_{cmd}}+a+b).$$ Then, the evolution of the particles is simulated using the GP models, thus obtaining the correspondent landing position $\p_T^{(m,i)}$. 
Then, we defined the following objective function $F(a,b, D)$, which measures the distance between the actual landing position and the simulated one,
\begin{equation*}
    F(a,b, D) = \frac{1}{N}\sum_{i=1}^{N} \frac{1}{M_d} \sum_{m=1}^{M_d} \norm{\p_T^{(m,i)}-\p^i}
\end{equation*}

Then, the candidate parameters of the $t_d$ distribution are the solution of the optimization problem
\begin{equation}
\begin{split}
    (a^*,b^*) =& \text{argmin}_{a,b} F(a,b, D)\\
    & \text{s.t. } \tilde{t}_r^{(m,i)} \sim U(t_{r_{cmd}} + a, t_{r_{cmd}}+a+b) \\
    &\,\,\,\,\,\,\,\begin{cases}
        x_0^{(m, i)} = h(\Pb_i, \vel_i, \tilde{t}^{(m,i)}) \\
        \x_{\tp+1}^{(m, i)} \sim \mathcal{N} (\mu_{\tp+1}^{(m, i)}, \Sigma_{\tp+1}^{(m, i)})
    \end{cases}
\end{split}
    \label{eq:delay_dist_opt_problem}
\end{equation}

\begin{figure*}[h!]
    \centering
    \includegraphics[width=\textwidth]{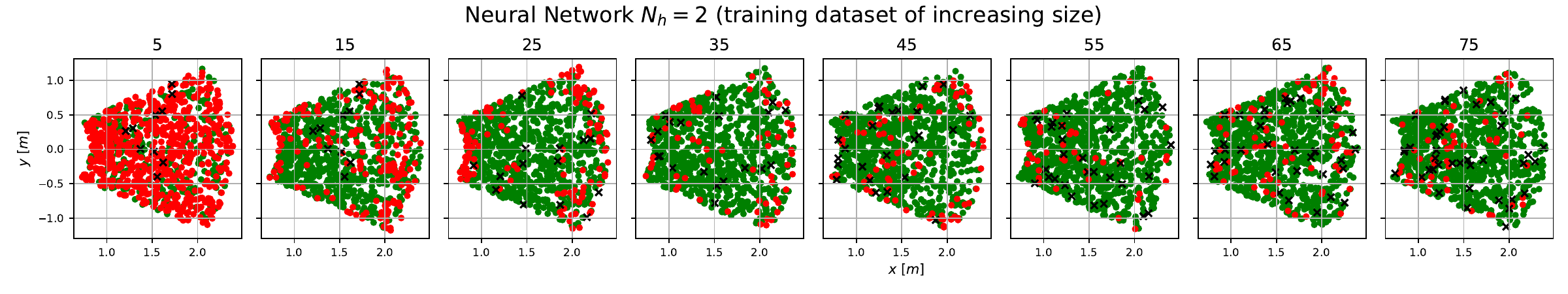}
    \caption{Results of the Neural Network policy trained on datasets of increasing size. 
    The target positions projected on the horizontal plane are colored green if the target was reached, and red if instead it was missed. The black markers represent the points of the training datasets.}
    \label{fig:unsupervised_learning_sim}
\end{figure*}

\begin{figure}[h!]
    \centering
    \subfloat[]{\includegraphics[width=\columnwidth]{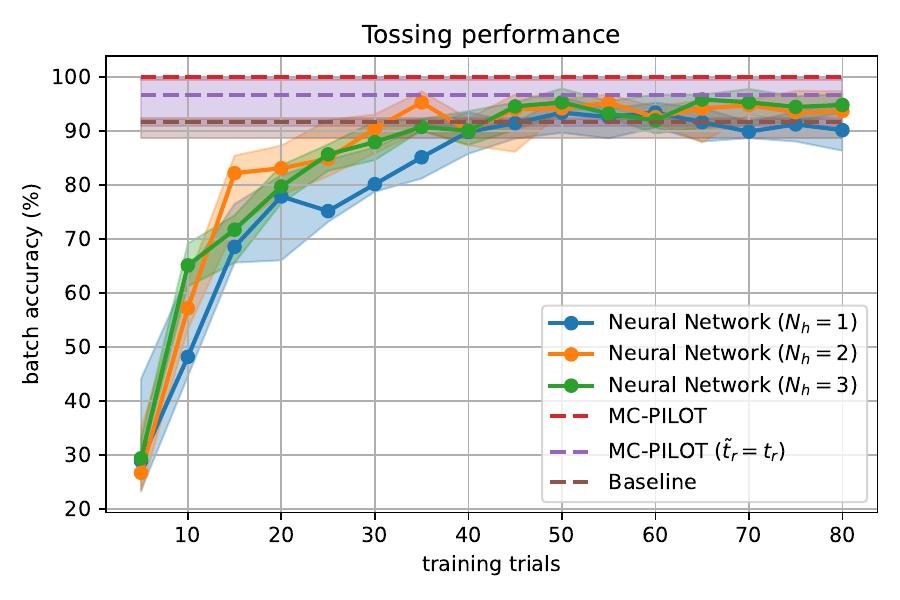}}

    \subfloat[]{\includegraphics[width=0.35\columnwidth]{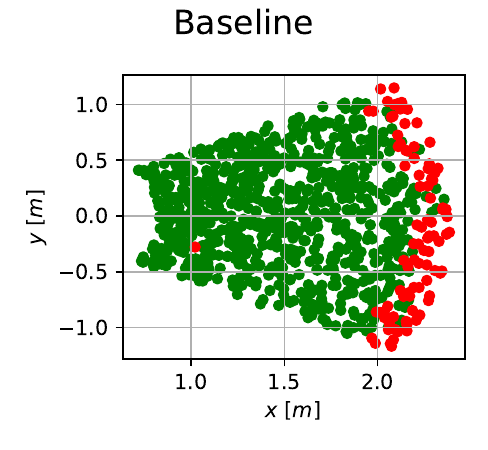}}
    \subfloat[]{\includegraphics[width=0.325\columnwidth]{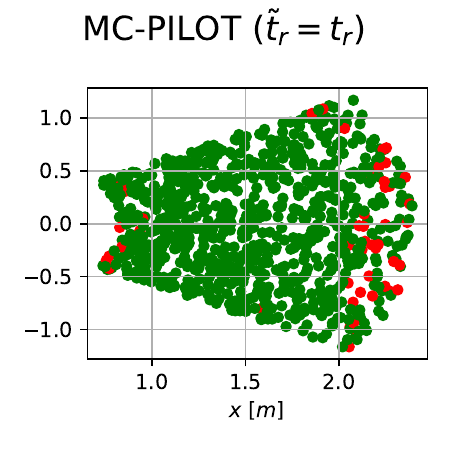}}
    \subfloat[]{\includegraphics[width=0.325\columnwidth]{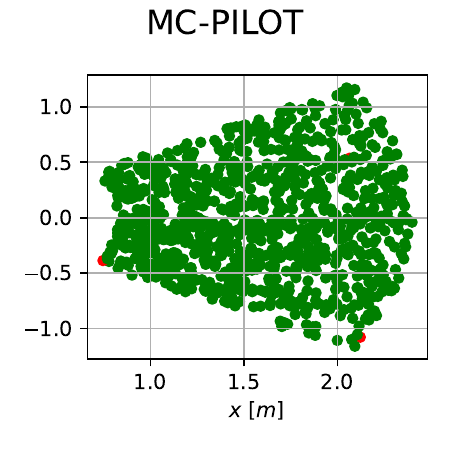}}
    \caption{(a): Plots of the target reach accuracy of 10 batches of trials for each tested policy, showing the median, and the first and third quartiles. \\
    (b-d): Results of tests with the Baseline and the MC-PILOT policies, the target positions projected on the horizontal plane are colored green if the target was reached, and red if instead it was missed.}
    \label{fig:policies_sim}
\end{figure}

The minimization problem in \cref{eq:delay_dist_opt_problem} can be solved using any search algorithm, in this work, we solve the optimization problem using Bayesian Optimization (BO)\cite{frazier2018tutorial_bayesian_opt, review_bayesian_optimization}. BO builds a surrogate for the objective function, by fitting a GP to input-output examples of the target fitness function and searches for the global minimum of the function in the posterior distribution of the GP. The posterior distribution is computed as described in \Cref{sec:gp}, and the search of the global minima can be done using gradient-based or gradient-free techniques. 
The global search algorithm comprises two parts: (i) a uniform random initialization and (ii) a guided search in the posterior distribution of the GP.
The first step uniformly samples several points from the domain of the objective function, to initialize the training dataset.
Then in the second part, an iterative procedure samples new points from the domain of the fitness function using an acquisition function, i.e. a function that selects a point $(a^+, b^+) $ in the function's domain using certain criteria. In this case, we used the Upper Confidence Bound (UCB), namely, at each iteration the procedure samples a new point with this rule:
\begin{equation}
    (a^+, b^+) = argmin_{a,b} \E[\hat{F}(a,b)] - \sigma \cdot \sqrt{var[\hat{F}(a,b)]},
    \label{eq:bo_ucb}
\end{equation}
which regulates the exploration-exploitation ratio by the parameter $\sigma$.
After each sampling, the point is evaluated, and the pair $[(a^+, b^+), F(a^+, b^+)]$ is added to the GP's training dataset. The procedure terminates once a maximum number of iterations is reached.

The output of the procedure is an estimated delay distribution $t_d \sim U(\hat{a},\hat{a}+\hat{b})$, where $(\hat{a}, \hat{b})$ is the sampled point that yielded the best objective. If $\hat{a}$ is not close to zero, then the system presents a residual minimum delay that is not compensated. We will see in \Cref{subsec:real_setup} that this is the case for all the tested objects, as mentioned in \Cref{subsec:motion_plan}. The $t_d$ distribution can be directly used in policy optimization as explained in \Cref{sec:mc_pilco_obj_throwing}. 
However, if $\hat{a}$ is too high, the algorithm can not learn an effective policy since performance is intrinsically limited by the delay offset, and the policy acts just on the release velocity. For this reason, we recompute the release command time to $t_{r_{cmd}} = t_r - \hat{a}$, to use both in policy optimization and in testing. 

\urldef\frankahandurl\url{https://download.franka.de/documents/220010_Product%20Manual_Franka%20Hand_1.2_EN.pdf}

\section{Experimental results}
\label{sec:exp}
This section describes the experimental results of our proposed algorithm for object-throwing both on simulated and actual setups, using the Franka Emika Panda robot. The Panda robot is a 7 DoF collaborative manipulator, equipped with the Franka Hand \footnote{\frankahandurl} as a gripper. In the actual setup, we use a vision system to track the object's geometrical centers and estimate their trajectories. 
The vision system is based on a camera network of 4 sensors (i.e., Intel Realsense Depth Camera D455), rigidly fixed to walls and positioned to observe the space around the robot. An object tracking algorithm analyses the RGB multi-view images provided by the cameras. 
The entire camera network was calibrated through hand-eye calibration techniques \cite{calibration_1, calibration_2}, to express the pose of each camera w.r.t. the robot base and to optimize the reciprocal pose between the different cameras.
As in most commercial systems, the Franka Emika gripper is not controllable in real-time and the robot and the gripper control are executed on separate threads. This means that it is not possible to obtain perfect synchronization between joints and gripper trajectories. Moreover, the gripper does not allow for measurement of the opening time, as the motor encoder readings are not available with reliable frequency. 
Due to the robot's stringent joints' velocity and acceleration limits, a throwing motion like the one presented in \cite{tossingbot} is not practical, as it results in a low maximum cartesian velocity, around $1.2$ $[m/s]$. Instead, we implemented a throwing motion similar to \cite{robot_skill_learning_deep_autoencoder_ball_throwing}, where the robot picks the object in front of itself and moves three joints to throw it behind itself. Therefore in the release configuration, the Panda robot is almost completely extended upwards, reaching a higher maximum cartesian velocity. A side effect of this strategy spawns when throwing at low speed. Indeed, the standard Franka Emika gripper has short fingers, causing picked objects to touch the gripper's body when the robot moves at low speed.
For this reason, we developed custom 3D-printed prosthetic fingers that prevent contact. The dimensions of these prosthetic tool tips are considered in the calculations for forward and differential kinematics. The prosthesis is visible in \cref{fig:robot_pic}, while the upper subfig. of \cref{fig:panda_sim} shows the robot in the release configuration.

To validate our approach, we first tested the motion planning and the MBRL algorithm MC-PILOT on a simulated setup. Screenshots of the simulated environment are reported in \cref{fig:panda_sim}.

All machine learning algorithms were implemented using the \emph{Pytorch} library \cite{pytorch}. For the implementation of BO we used the python library \emph{bayesian-optimization} \cite{bayesian_opt_library}. \Cref{table_hyperparam} reports the values of the main parameters of MC-PILOT used in the simulation and the real setup.

In the next subsections, we first describe the object-throwing system implemented with the Panda robot, including details of the motion planning algorithm, second, we present the results of the object-throwing task in simulation, and finally the results on the actual system.

\begin{figure}
    \centering
    \includegraphics[width=0.8\columnwidth]{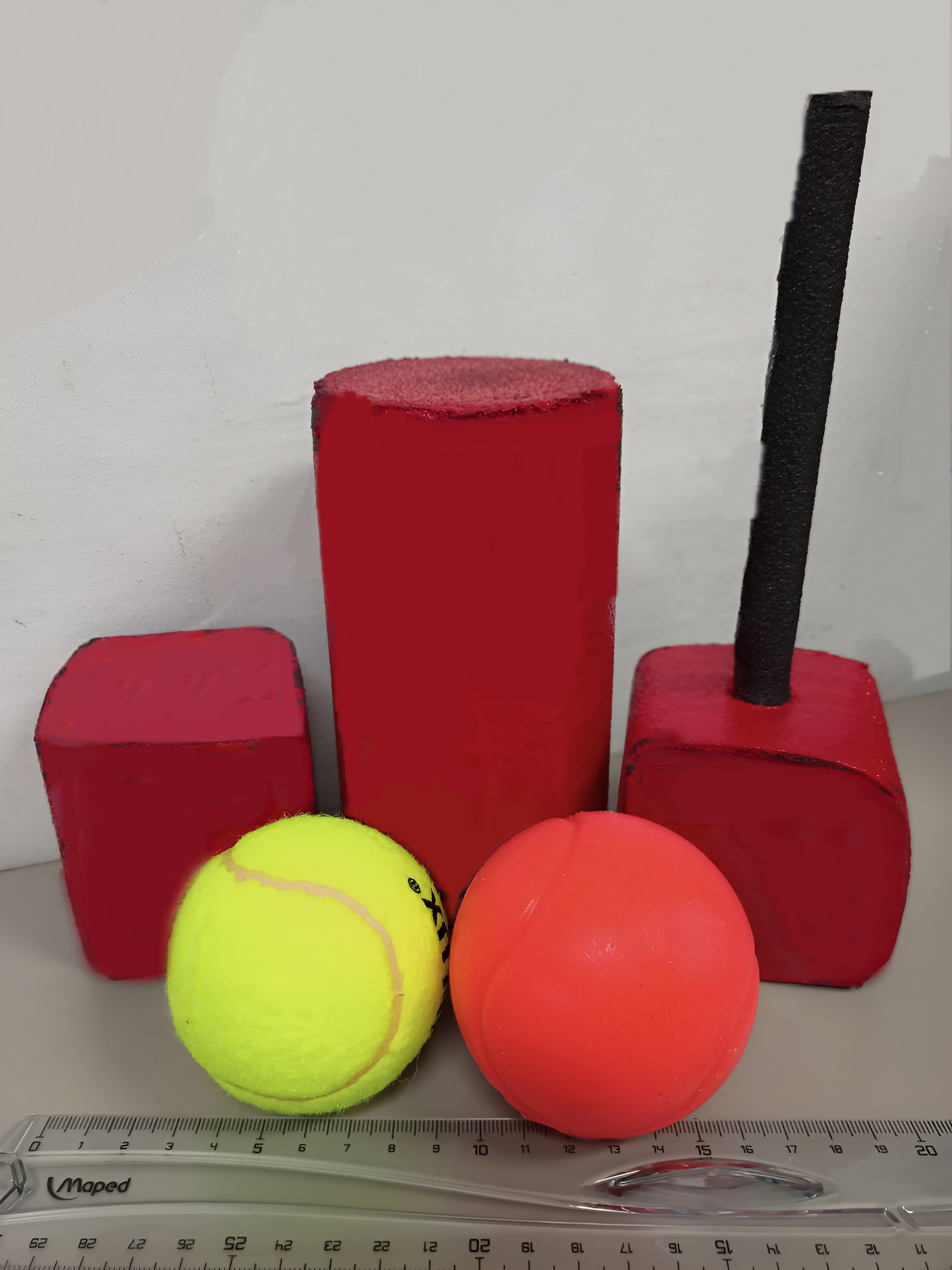}
    \caption{The tested objects. On the front row, from left: the tennis ball, and the rubber ball. On the back row, from left: the cube, the cylinder, and the hammer}
    \label{fig:objects}
\end{figure}

\subsection{Robot trajectory generation}
\label{subsec:deterministic_trajectory}
This section details the implementation of an object-throwing system using the Franka Emika Panda robot equipped with the Franka Emika gripper. 
The approach can be adapted to all articulated industrial manipulators. For all the planning done in joint space, we referred to the Denavit-Hanterberg convention and specifications reported on the manufacturer's website \footnote{\url{https://frankaemika.github.io/docs/control_parameters.html}}.
The tossing motion is achieved by defining an appropriate joint position and velocity profile, compliant with the description of \Cref{subsec:motion_plan}. This approach is very similar to other works, for example, in \cite{tossingbot} the robot's motion is defined by motion primitives, and the throwing policy regulates the throwing velocity.

The following robot configuration was chosen as the release configuration:
\begin{equation}    
    \q_{rel}=\begin{bmatrix}\gamma & -\frac{25}{180} \pi & 0 & -\frac{\pi}{4} & 0 & \pi & 0\end{bmatrix}^T,
\end{equation}
where $\gamma$ is the position of the first joint, which is set to align the robot motion with the target point.
The value of the second joint is negative due to the chosen convention. The upper subfig. of \cref{fig:panda_sim} shows the robot in the release configuration. 
The nominal release point can be described as in \cref{eq:release_point}, with $\ell_r=0.07$ $[m]$ and $z_{rel}=1.50$ $[m]$.
To describe the relation between the different joint velocities at release, we introduce the vector:
\[\dq^* = \begin{bmatrix}0 & -1 & 0 & 1 & 0 & 1 & 0\end{bmatrix}^T\]
i.e. joints 1,3,5 and 7 are motionless, while the other joints must reach the same speed at release.
Then it is possible to compute the vector $\velbf^*$ as
\begin{equation}
    \velbf^* = J_a(\q_{rel}) \dq^*,
\end{equation}
which is such that $\frac{\velbf^*}{\norm{\velbf^*}} = \velbf_\gamma^\alpha$, with $\alpha \sim 0$.
We can define a $\dq_{rel}$ vector, as function of the desired release velocity $\vel$ as
\begin{equation}
    \dq_{rel}^\vel = \frac{\dq^*}{||\velbf^* ||} \vel,
\end{equation}
and the resulting cartesian velocity $\velbf_{rel}$ can be computed as:
\begin{equation}
    \velbf_{rel} = J_a(\q_{rel}) \dq_{rel}^\vel = J_a(\q_{rel})  \frac{\dq^*}{||\velbf^* ||} \vel = \vel \frac{\velbf^*}{||\velbf^* ||} = \vel \velbf_\gamma^\alpha,
\end{equation}
consequently, $\velbf_{rel} \parallel \velbf_\gamma^\alpha$ and $\norm{\velbf_{rel}} = \vel$.

Polynomial profiles $\dq_{t}^\vel$ and $\q_{t}^\vel$ are designed to comply with the requirements described in \Cref{subsec:motion_plan}, i.e.
$\q_{t_r}^\vel = \q_{rel}$ and $\dq_{t}^\vel = \dq_{rel}^\vel$.
On the real manipulator, a joint velocity controller provided by the manufacturer's robot control library \emph{libfranka}\footnote{\url{https://github.com/frankaemika/libfranka}} is used to move the robot. The profiles are generated at a frequency of \SI{1000}{\Hz}. \Cref{fig:reference_trajectories} presents the joints reference trajectories for the 3 joints responsible for the motion, related to the minimum and maximum cartesian velocities of the system. Both trajectories are such that the release configuration is reached when the velocities are at their peak, to have the desired cartesian velocity at the end-effector. \Cref{fig:throwing_motion} depicts the robot motion.

\subsection{Simulation}
The simulated environment was implemented in Gazebo \cite{koenig2004design_gazebo} and the ROS framework \cite{koubaa2017robot_operating_system_ros}, using the robot's model and simulation libraries provided by the manufacturer. 
In the experiments, we use as a target a hollow cylinder of height $0.1$ $[m]$, radius $0.05$ $[m]$, and border thickness $0.01$ $[m]$. As bullet, we used a sphere of dimensions of a standard golf ball, namely, a radius of $0.0215$ $[m]$ and mass of $0.02$ $[Kg]$. 
\begin{figure}[h!]
    \centering
    \includegraphics[width=\columnwidth]{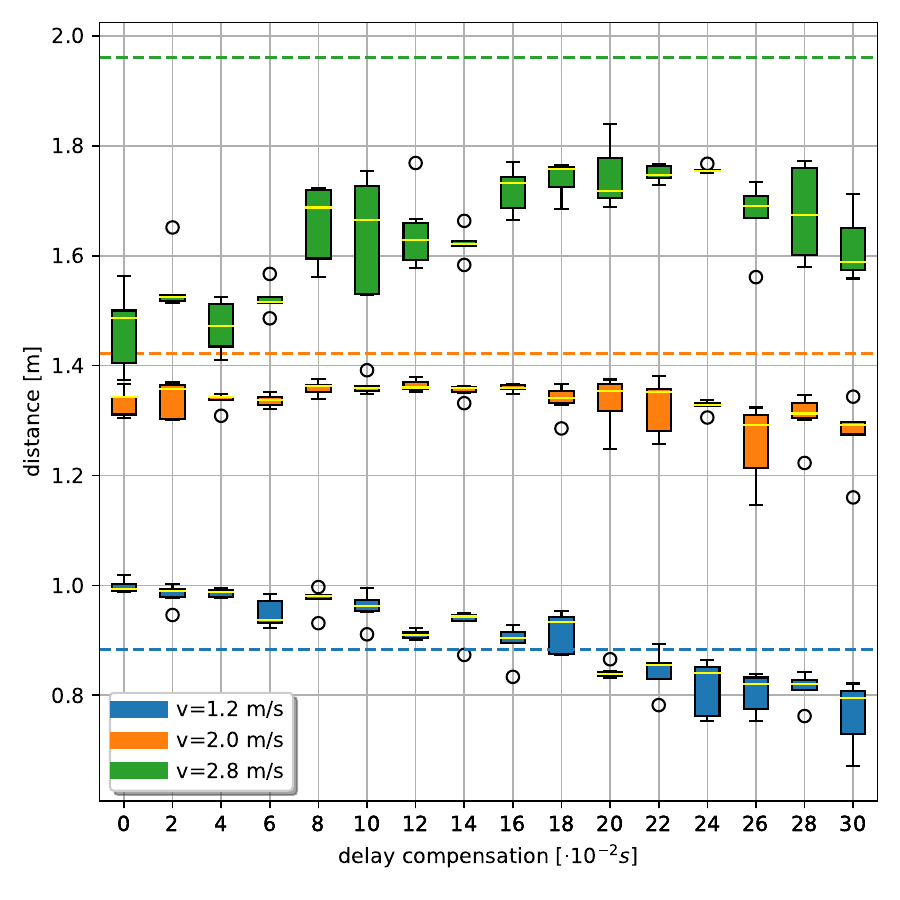}
    \caption{Delay compensation estimation experiment with the rubber ball. For each release velocity, we plot the landing distance distribution as a function of delay compensation. Dashed lines are the nominal landing distance computed with the ballistic equations.}
    \label{fig:delay_compensation_estimation_distances}
\end{figure}
We implemented the air drag effect of flow around a sphere following the drag equation of fluid dynamics:
\begin{equation}
    \Fb_D = -\frac{1}{2} \rho C_D(\dot{\p}_{t}) A \norm{\dot{\p}_{t}}\dot{\p}_{t}
\end{equation}
where $F_D$ is the drag force applied on the object moving in the fluid, $\rho$ is the density of the fluid, $A$ is the cross-sectional area of the object, $C_D$ is the drag coefficient, which depends on the \textit{Reynolds} number $R_e(\dot{\p}_t) = \frac{\norm{\dot{\p}_t} 2 r}{\nu}$. $r$ is the sphere's radius and $\nu$ is the kinematic viscosity of air. Since $C_D$ depends on the fluid velocity, to compute it for each time instant of the simulation, we used the empirical relationship of the drag coefficient to the fluid speed derived in \cite{ALMEDEIJ2008218_drag}.
To simulate the release errors of the manipulator, the objects are released with a delay uniformly sampled in the interval $[\SI{0.01}{\s}, \SI{0.02}{\s}]$. In all simulated experiments we set $t_{r_{cmd}} = t_r$. The \textit{MoveIt!} \cite{coleman2014reducing} package provides the robot interface, and, to control the execution of the tossing trajectory, we used an effort joint trajectory controller supplied by the \textit{ros\_control} framework \cite{ros_control}. 

In the experiments, we considered a setting similar to the real-world setup, the robot's base is at a certain height w.r.t. the ground, and the target cylinder is moved arbitrarily in the space at ground level, as depicted in \cref{fig:panda_sim}. 
Since the goal of the task is to throw the bullet into the cylinder, we set the $x_P,y_P$ coordinates of the target position to be the geometrical center of the cylinder's base, while the $z_P$ coordinate is the height of the cylinder.

We apply MC-PILOT to the simulated environment with $N_{exp}=5, N_a=0$, which is sufficient to yield a performing policy to solve the task in this setting, with a single execution of the algorithm's outer loop (1 trial). To obtain a performance statistic, we test the algorithm on $10$ different random seeds, for each seed the policy is tested on the system by sampling a batch of $100$ target locations. For each batch the accuracy of the policy w.r.t. the task definition is computed as the ratio between correct and total throws.

The delay optimization procedure described in \Cref{sec:delay_dist_opt} is used to estimate a uniform distribution $U(\hat{a}, \hat{a}+\hat{b})$ which is used to sample the delay to initialize the particles with \cref{eq:particles_init_noise}. In the BO, we limit the search domain of $a$ and $b$ respectively to $[-0.3, 0.3]$ and $[0, 0.01]$. In \cref{table_opt_delays} (left) we report the mean and standard deviation of the delay distribution parameters estimated by the optimization procedure in \Cref{sec:delay_dist_opt} on the simulated data of the 10 different seeds. We can see that even if the estimated distributions are different from the real delays, the optimization does predict the presence of a delay statistically, as the mean of $\hat{a}$ is marginal $\geq 0$.

To highlight the effect of modeling the system uncertainties, we also test a policy trained with MC-PILOT, without modeling the release delay, i.e. with $\tilde{t}_r = t_r$, using \cref{eq:noiseless_particles_init} to initialize particles. We refer to this policy as MC-PILOT ($\tilde{t}_r = t_r$).

As references, we also present the performance collected in the same setting by the Baseline policy from \Cref{subsec:baseline_policy} and policies obtained with a Model-Free method based on neural networks. 
A Neural Network policy was obtained by training a standard feedforward network model for the object-throwing task.
The Network's input is a vector $\Pb \in \R^3$, representing a target point, while its output is a single positive scalar, representing the input velocity.
A training dataset for the Neural Network is collected by throwing the ball in random directions with random input velocities. The recorded landing positions and input velocities compose the training set of the Network, the landings are the inputs of the Network, while the applied velocities are the targets. The network is then trained to solve the regression problem, to minimize the mean squared error on the training set, with a standard gradient approach using the optimizer Adam \cite{kingma2014adam}. We tested three Neural Network architectures for the object-throwing task. The architectures share a common backbone composed of $N_h$ feed-forward hidden layers of 200 neurons, with \emph{ReLU} activation function and a final output layer with a squashing activation function defined as
\begin{equation}
    g(\cdot) =  \frac{u_{M}}{2} \left(  \tanh{ \left(  \cdot  \right)} +1 \right),
\end{equation}
similar to the output of the policy \cref{eq:tossing_policy}.
The three architectures differ in the number of hidden layers $N_h$, we tested architectures with $N_h=1, 2, 3$.

Results are summarized in \cref{fig:unsupervised_learning_sim,fig:policies_sim}. 
\Cref{fig:policies_sim}-(a) compares accuracies of the Baseline, Neural Network policies, and MC-PILOT with and without delay estimation. The statistics of the accuracy level of MC-PILOT and the Baseline are compared with the statistics of the Neural Network policies as their training dataset grows. \Cref{fig:policies_sim}-(b-d) visualize how the Baseline and the MC-PILOT policies perform the task, by showing the distribution of the correct and failed throws. The same is done in \cref{fig:unsupervised_learning_sim} for the Neural Network policy with $N_h=2$ hidden layers, for increasing dimensions of the training dataset.

\Cref{fig:policies_sim}-(a) shows that in the considered setup, the MC-PILOT algorithm yields a policy with almost 100\% percent accuracy, overcoming the limitations of the Baseline. \Cref{fig:policies_sim}-(d) shows that all target positions are reached with strong repeatability.
If in the same setup, with the same algorithm, we do not model the uncertainties due to delays, the accuracy degrades. This is visible in \Cref{fig:policies_sim}-(c) where we can see that the policy fails to hit some of the further targets, similarly to the Baseline, as shown in \Cref{fig:policies_sim}-(b). Still, the policy trained by MC-PILOT shows higher accuracy than the baseline, even if the release uncertainties are ignored, due to the GP model capturing the drag effects.
Finally, as reported in \cref{fig:policies_sim}-(a), the Neural Network policies indeed increase the accuracy level as their training dataset grows in size, but ultimately cannot reach perfect accuracy in the tested setup. The 3 considered Neural Network architectures perform similarly as the training dataset dimension grows. For the bigger dimensions tested, it seems that the Networks with $N_h=2,3$ perform better than the Networks with $N_h=1$, stabilizing to a performance level comparable to MC-PILOT ($\tilde{t}_r = t_r$). In \cref{fig:unsupervised_learning_sim} one can see that as the training dataset's coverage increases, the number of reached targets also increases. The distribution of target misses does not have a particular shape, different from the Baseline and MC-PILOT ($\tilde{t}_r = t_r$), which might be only related to the richness of the training dataset.

\subsection{Actual Setup}
\label{subsec:real_setup}
We tested the MC-PILOT algorithm on the actual setup, described at the beginning of this section, using three different objects: a rubber ball, a tennis ball, and a cube. Moreover, two additional objects of more complex shapes are also used. A picture of the objects is reported in \cref{fig:objects}.
In our setup, the robot is mounted at a certain height w.r.t. the ground, and the target space is at ground level, behind the robot. 
For convenience, in the first experiments with the real setup, we do not use a bin or a box as targets. Targets are just cartesian coordinates on the ground. A target hit or miss is evaluated by comparing the target position and the landing position. Instead, in the final experiment,  discussed at the end of \Cref{subsubsec:final_experiments}, we used a bin as a target. Interestingly, MC-PILOT derived the throwing policy for the bin from data collected in the first experiments, thus confirming the great adaptability of our approach. 

The rest of this section is organized to highlight the main achievements. First, we provide an evaluation of the setup, including an initial approximation of the gripper opening command time and an assessment of the target space dimension. Next, we evaluate the effectiveness of the GPR model learning procedure described in \Cref{subsec:mc_pilot_model}, comparing it to both real data and the nominal model based on ballistic equations. Finally, we present the results of applying MC-PILOT to the PnT task.

\begin{figure}
    \centering
    \subfloat{\includegraphics[width=\columnwidth]{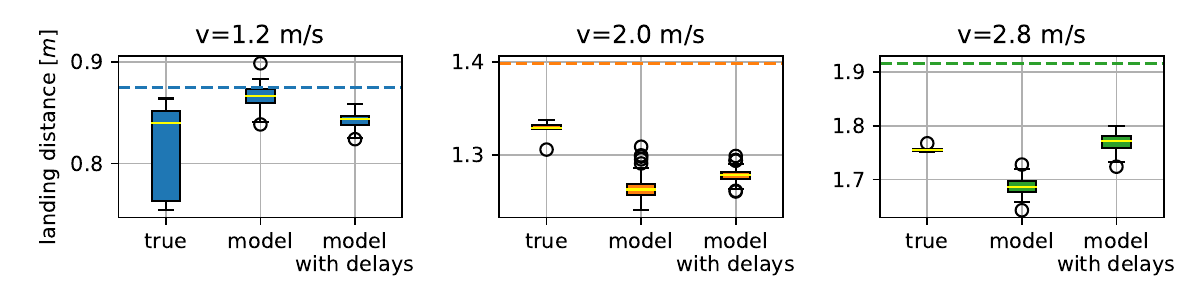}}
    

    \subfloat{\includegraphics[width=\columnwidth]{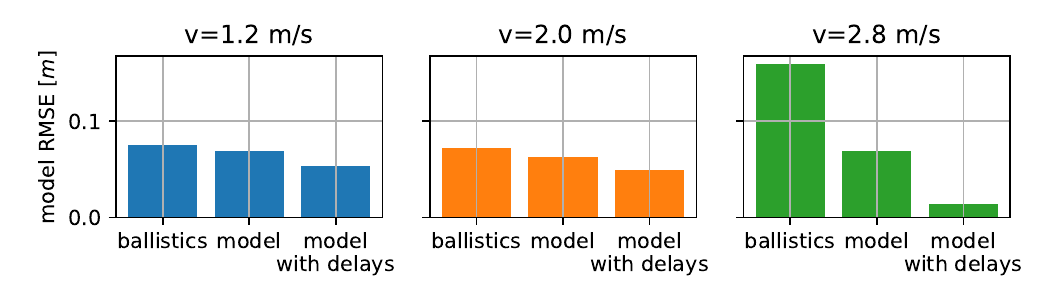}}

    \caption{Top: landing distances predicted by the the simulations with the GP model, and the ballistic equations predictions. Bottom: prediction errors of the same models.}
    \label{fig:test_model}
\end{figure}

\begin{figure}[t]
    \centering
    \includegraphics[width=\columnwidth]{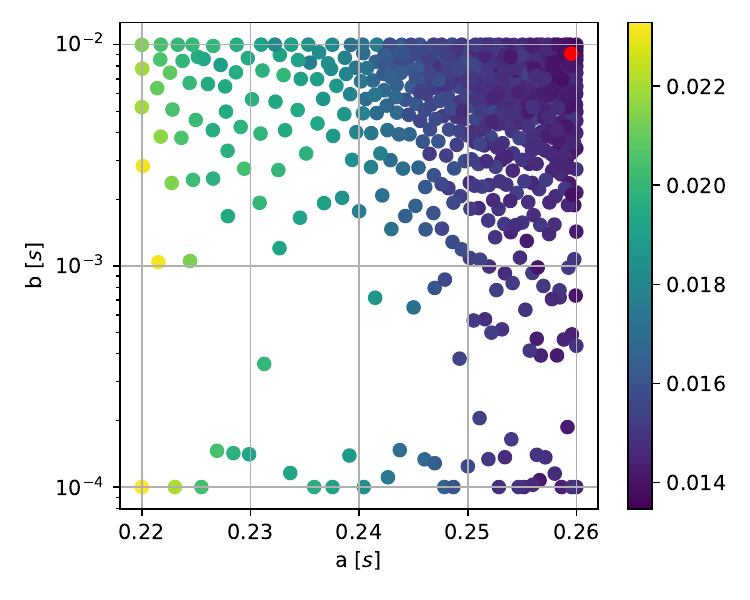}
    \caption{The points sampled in the delay distribution optimization of the rubber ball. Points are colored following the color bar on the right, based on the related objective function value. The red point is the best sample found.}
    \label{fig:delay_dist_opt}
\end{figure}

\subsubsection{Setup Evaluation}
\label{subsubsec:setup_evaluation}
As previous works have shown \cite{softToss_gripper}, for a correct release of the bullet in mid-motion, it is necessary to account for the gripper opening delay, i.e. the time necessary for the grasped objects to detach from the tooltips. 
We set up an experiment to roughly estimate the opening delay, by throwing an example object, a rubber ball, at three different velocities: $1.2$, $2.0$, $2.8$ $[m/s]$, where the latter is the maximum achievable velocity on the real setup. For each of these velocities, we throw the ball multiple times with delay compensation values $t_r - t_{r_{cmd}}$ between $0$ and $30 \cdot 10^{-2}$ $[s]$ and measure the horizontal distance traveled by the ball before landing. This experiment gives the statistics reported in \cref{fig:delay_compensation_estimation_distances}. 
The figure reports, for each tested value of $t_r - t_{r_{cmd}}$, and for each tested throwing velocity, the statistics of the ball's travel distance, estimated by measuring the distance on the horizontal plane between the origin of the robot's reference frame and the measured point of collision with the ground. In the same figure, we present the nominal landing distance calculated using the ballistic equations for reference.
The high variability between traveled distances obtained with different delay compensations highlights that a correct handle of this aspect is crucial to obtain an effective PnT algorithm.  
As a first approximation, the selected delay compensation is $24 \cdot 10^{-2}$ $[s]$, which gives results closest to the nominal travel distance computed with the ballistic equations. 
It must be noted that this is a rough estimation done with an example object, and due to the elasticity of the prosthetic tooltips, results vary when using different objects. Indeed, a residual delay is likely to be still present in the system with the selected compensation value.
This estimated delay compensation value is used for the Baseline policy, with all objects, namely we set the opening command time to $t_{r_{cmd}}=t_r-24 \cdot 10^{-2}$ $[s]$.
Notice that once selected $t_{r_{cmd}}$, we also get an estimate of the reachable target space dimensions, given by the landing distances related to the top and lowest release velocity.
\begin{remark}
Note that this first rough estimation is required for all learning-based and analytical methods, notice that in \Cref{fig:delay_compensation_estimation_distances} it is clear that no compensation of delay results in very limited target domain dimensions.
\end{remark}

\begin{figure}
    \centering
    \subfloat{
    \includegraphics[width=\columnwidth]{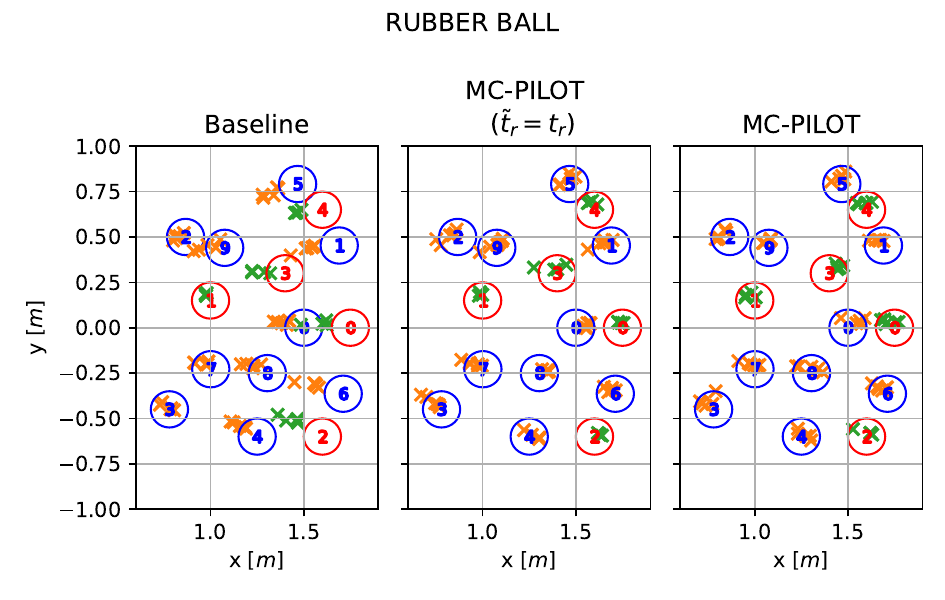}}
    
    \subfloat{\includegraphics[width=\columnwidth]{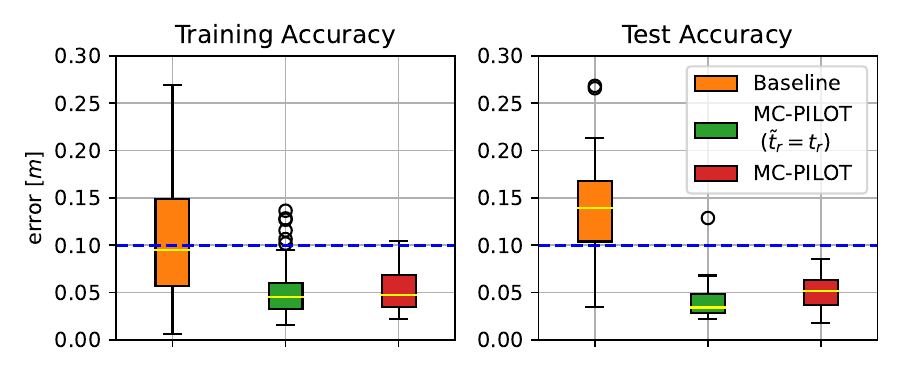}}
    
    \caption{Accuracy tests with the rubber ball. 
    The numbered circles are placed in the training (blue) and test (red) positions, their radius is $\SI{0.1}{\m}$. The crosses are the landing recorded positions.
    The boxplots quantitatively compare the policies' accuracy on the training and test targets.}
    \label{fig:results_rubber_ball}
\end{figure}

\begin{figure}
    \centering
    \subfloat{\includegraphics[width=0.75\columnwidth]{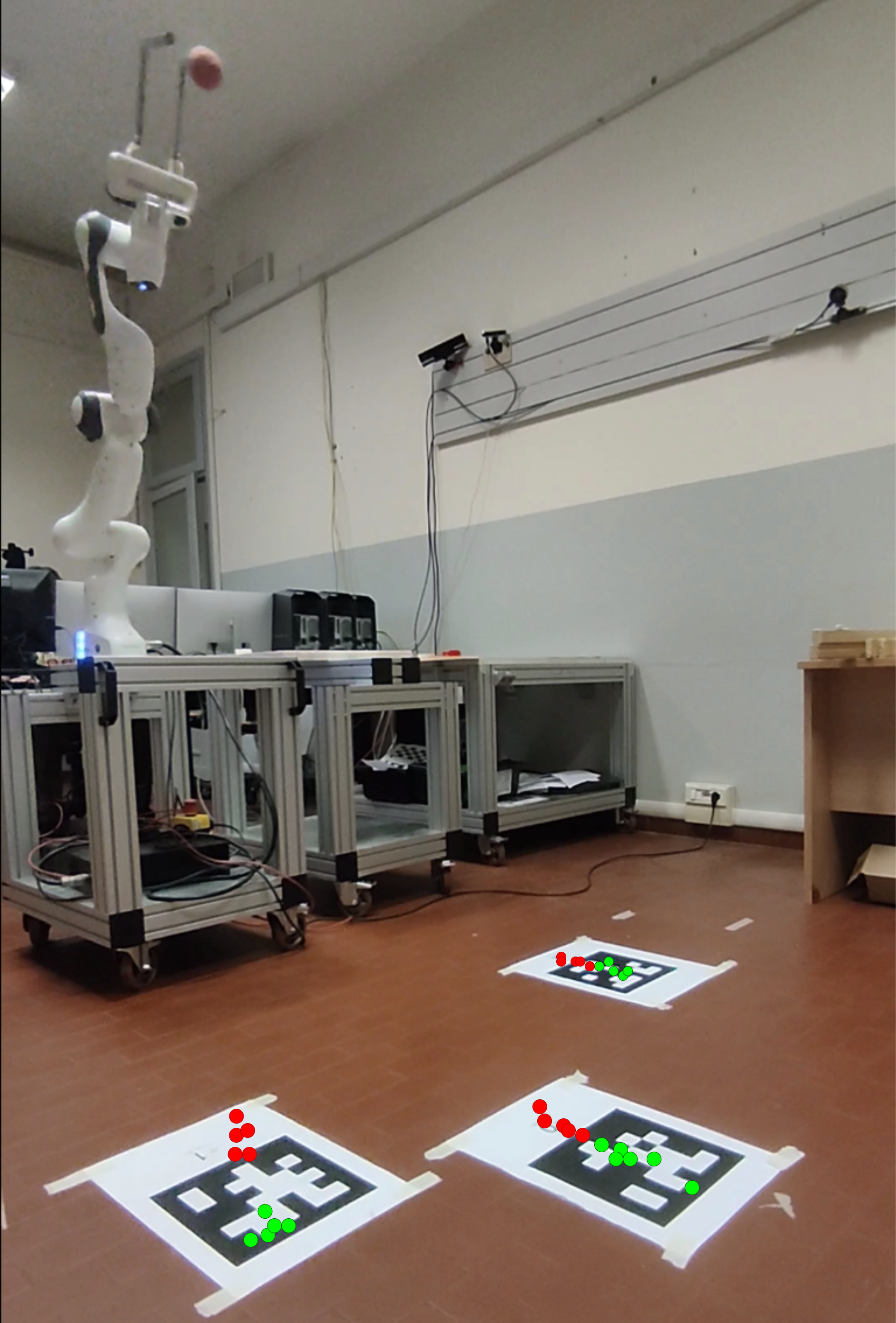}}
    \caption{A picture of the robot in the setup, executing the task with the rubber ball. The April tags' centers are the demo targets, the green points are the landing positions obtained with MC-PILOT, the red points are obtained with the Baseline.}
    \label{fig:rubber_ball_demo}
\end{figure}

\subsubsection{Model Learning}
\label{subsubsec:model_learning_real_setup}
The model described in \Cref{subsec:mc_pilot_model} was first validated using the data from the experiment in \cref{fig:delay_compensation_estimation_distances}. The trajectories obtained by throwing the rubber ball with the opening command time set to $t_{r_{cmd}} = t_r - 24 \cdot 10^{-2}$ $[s]$ are used to train a GP model. 
With the model we then simulate a batch of particles for each of the three cartesian velocities, the initialization is performed with \cref{eq:noiseless_particles_init}, simulating the system in ideal conditions, and with \cref{eq:particles_init_noise} using the delay estimated with the procedure described in \cref{sec:delay_dist_opt}. 
For each of the three tested release velocities, 100 particles are initialized with base joints position $\gamma = 0$.
Results of the simulated particles are reported in \cref{fig:test_model}. The top plot reports the landing distances of the collected trajectories, the GP model predictions, with and without the estimated delay as boxplots, and the travel distance computed with the ballistic equations as horizontal lines. Along with that, in the bottom plot of \cref{fig:test_model} we report the RMSE if the different models. For each model, we considered as a target the mean of the simulated landing positions, and we computed the RMSE of the actual landing positions; in the case of the balistic model the target is equal to the nominal landing distance obtained solving \cref{eq:balistic-no-frict}, with $d$ as unknown variable.
It is possible to notice that for the middle and top velocity, the projectile lands not as far as the ballistic equations predict, this is due mostly to the unmodeled air friction. Instead, the models' predictions are closer to the real data, particularly for the model with delays. 
We can see that the model without delays predicts a landing further than the real data for the low velocity, while for the higher velocities, it predicts shorter than real landings. This can be explained considering that the GP one-step model takes into account air friction, which is more present at higher velocities.
The modeling of delays, instead allows to account also for errors in the initial state of the projectile, including both position and velocity. Indeed, these release delays have two intertwining effects: they reduce the release velocity, and move forward the release position, thus creating the relations we see in \cref{fig:test_model}, since delays cause more errors when the robot moves faster.

\subsubsection{Throwing experiments}
\label{subsubsec:final_experiments}
First, we present results obtained with a rubber ball, and then we move to different objects to test the adaptability of the algorithm to variations of material, mass, and shape.

\emph{Rubber ball.} To deal with the complexity of the actual setup, in these experiments, we use $N_{exp}=10$ and $N_a=2$, which yield a performing policy with a single trial.
We applied the algorithm using the delay estimation of \Cref{sec:delay_dist_opt}, the left limit $\hat{a}$ is used to compute a new opening command time $t_{r_{cmd}} = t_r - \hat{a}$, and the delay is simulated as $t_d \sim U(\hat{a}, \hat{a} + \hat{b})$. In the BO procedure, we limit the search domain of $a$ and $b$ respectively to $[0.21, 0.27]$ and $[0, 0.01]$. In \cref{fig:delay_dist_opt} we report the results of the BO search procedure on the rubber ball.
The plot shows the sampled points color-coded based on the relative objective function value. Areas of low cost are more densely sampled, while the points are sparse where the objective is higher.

For the PnT task, training and test targets are randomly selected among the possible positions and kept constant throughout the experiments. Moreover, three additional target locations are obtained by measuring the position of April tags \cite{apriltags}, of $0.2$ $[m]$ size, manually placed on the target space. These three targets, visible in \Cref{fig:rubber_ball_demo}, are called demo targets hereafter.
As done in simulation, to showcase the effects of delay modeling we show the performance of a policy trained with MC-PILOT, with or without considering delay effects in the policy optimization (MC-PILOT $(\tilde{t}_r = t_r)$); in this case, as with the baseline, we set the opening command time to $t_{r_{cmd}}=t_r-24 \cdot 10^{-2}$ $[s]$, while the particles in the policy optimization are simulated with \cref{eq:noiseless_particles_init}. 
Qualitative and quantitative results are reported in \cref{fig:results_rubber_ball}. The results are presented both in the form of 2D plots, qualitatively comparing the performance of each policy, and as boxplots for a quantitative comparison. The 2D plots show the target locations as well as the landing positions, the latter are represented as circles of $0.1$ $m$ radius. We can consider a target hit if the landing is inside the $0.1$ $m$ radius. The boxplots instead show the statistics of the distance between the landings and the targets. 
Similarly to the experimental results obtained in simulation, the Baseline policy doesn't reach the furthest targets. MC-PILOT is instead capable of correctly throwing towards the training targets and the test targets, due to the high generalization of the Model-Based approach. MC-PILOT $(\tilde{t}_r = t_r)$ is an improvement over the Baseline, but it shows less robustness than MC-PILOT, indicating the importance of including control delays and uncertainties in the policy optimization. MC-PILOT $(\tilde{t}_r = t_r)$ makes mistakes in particular with the closer and further targets, this is because, as already shown in \cref{fig:test_model}, the GP model alone is not precise enough without modeling delays.

In \cref{fig:rubber_ball_demo} we show a demonstrative picture, showing the robot workspace and the demo target April tags. Green points on the ground are the landing positions obtained with MC-PILOT, while red points are the landing positions obtained with the Baseline.

\begin{figure}
    \centering
    \subfloat{\includegraphics[width=\columnwidth]{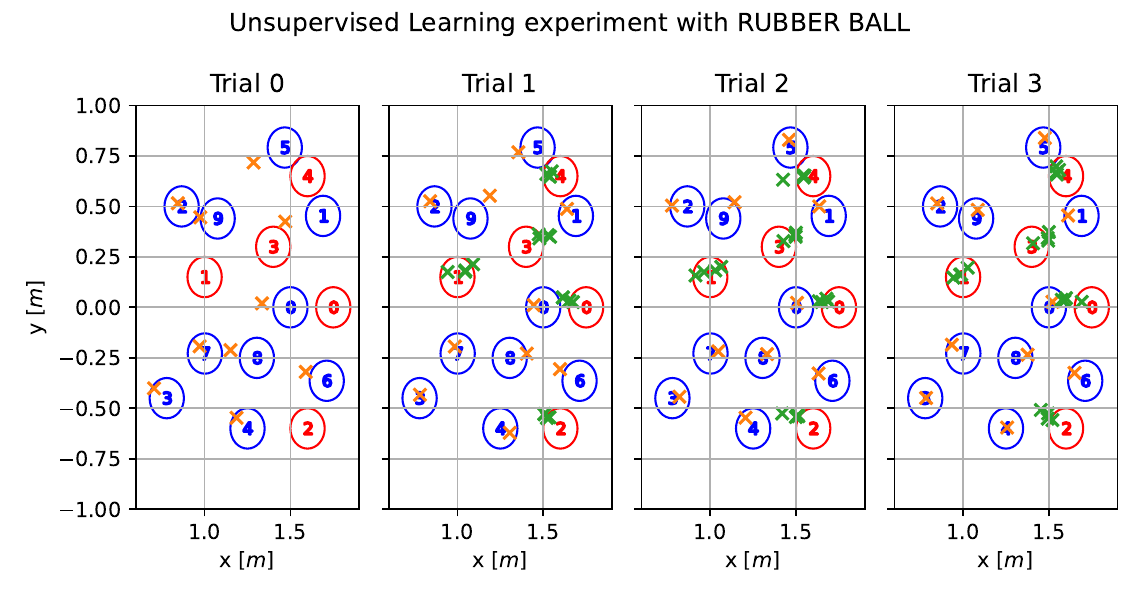}}

    \subfloat{\includegraphics[width=\columnwidth]{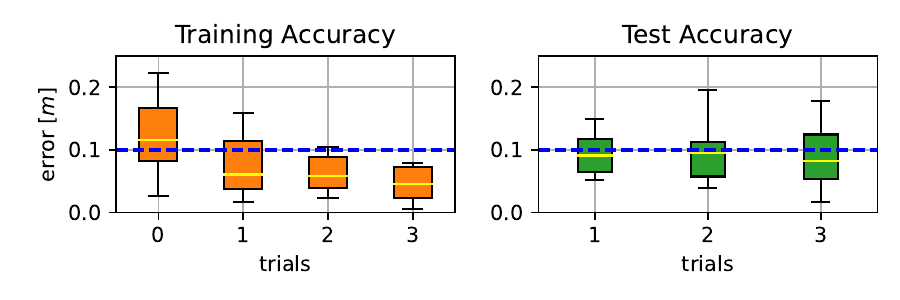}}
    \caption{Unsupervised learning experiment with the rubber ball.  The numbered circles are placed in the training (blue) and test (red) positions, their radius is $\SI{0.1}{\m}$. Each subplot shows the results of a trial, orange crosses are the landing positions of training targets, while the red crosses of test targets. The boxplots show the trend of accuracy over the training and test targets.}
    \label{fig:unsupervised_learning_rubber_ball}
\end{figure}

To compare our approach with a  model-free approach, we implemented also an unsupervised solution based on a Neural Network. We selected the Neural Network architecture that achieved the best results in the simulated experiments. 
The experiment is composed of successive trials. In trial 0, we build an initial training dataset for the Network, obtained by throwing the rubber ball toward the 10 training targets with the Baseline policy and recording the landing point. This initial dataset is used to train the Network to solve the regression problem, as done in the simulated experiments. 
In trial 1, the model's performance is evaluated by testing on the 5 test targets, for each test target we perform several throws and record the landing position. 
Then, the model is used to perform a throw toward each training target, recording the landing position. The data related to these 10 throws is added to the model's dataset, which is then used to retrain the model. The successive trials repeat the process of trial 1.
In \cref{fig:unsupervised_learning_rubber_ball} we report the results of 4 trials performed on the system with the unsupervised learning procedure. Indeed, by trial 3 the Neural Network has learned to perform the task on the training targets, but yet still fails some of the test targets. This suggests that the policy trained with unsupervised learning is not able to generalize to targets not seen during training, as seen in the simulated experiments. These results confirm that our model-based approach is much more data-efficient and more capable of generalizing to unseen targets. 

\begin{figure}
    \centering
    \subfloat{\includegraphics[width=\columnwidth]{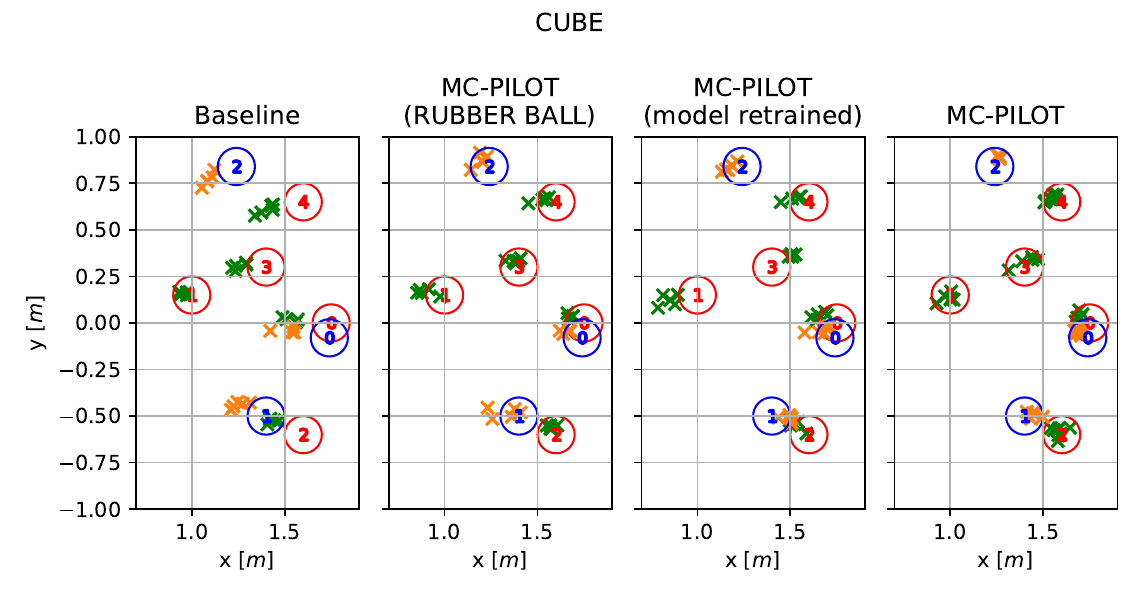}}

    \subfloat{\includegraphics[width=0.7\columnwidth]{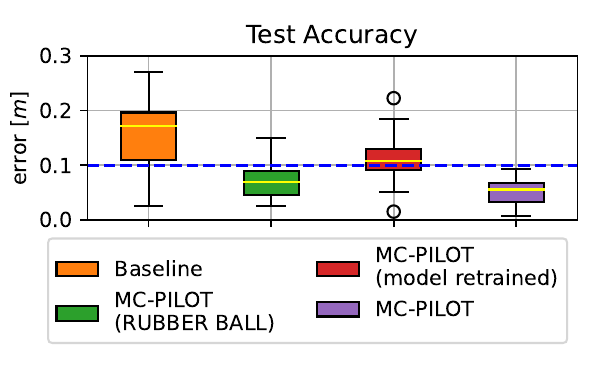}}

    \caption{Accuracy tests with the cube. The numbered red circles are test positions, while the blue circles are demo positions, their radius is $\SI{0.1}{\m}$.}
    \label{fig:results_cube}
\end{figure}

\begin{figure}
    \centering
    \subfloat{\includegraphics[width=\columnwidth]{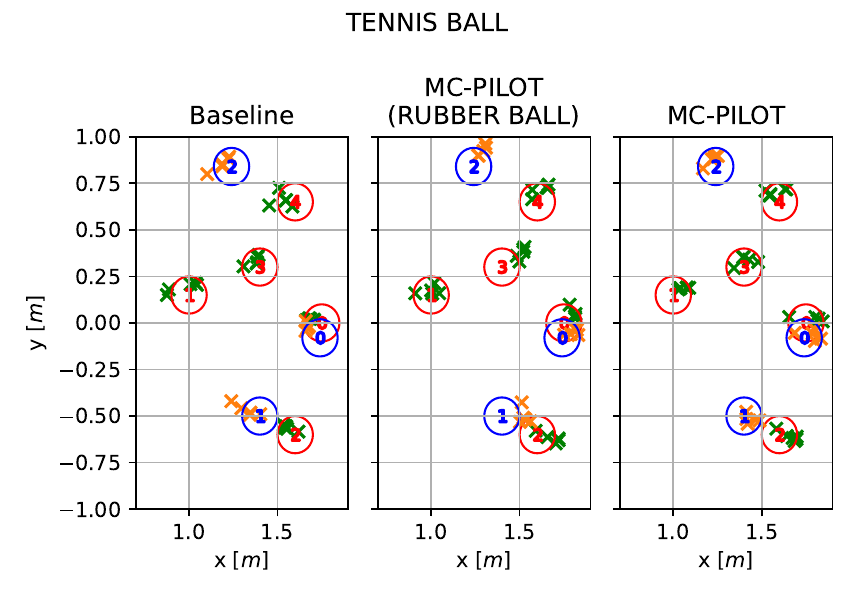}}
        
    \subfloat{\includegraphics[width=0.65\columnwidth]{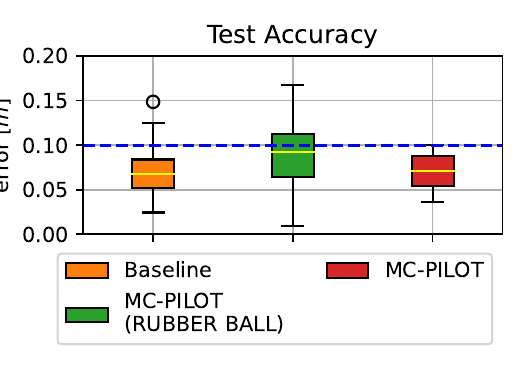}}
    \caption{Accuracy tests with the tennis ball. The numbered red circles are placed in the test positions, while the blue circles are placed in the demo positions, their radius is $\SI{0.1}{\m}$.}
    \label{fig:results_tennins_ball}
\end{figure}

\emph{Other objects.}
We tested our algorithm also using different objects, with different materials and shapes. First, we discuss results obtained with a tennis ball and a cube, reported in \cref{fig:results_cube}, and \cref{fig:results_tennins_ball}. In the figures, we report the results of the throws related to the testing targets and the three demo targets.
For both objects, we test the Baseline and the policy trained with MC-PILOT on the rubber ball and the policy trained with MC-PILOT and the delay estimation procedure on the object. For the delay estimation, we use the same search domain as for the rubber ball. For the cube, we also show the results of a policy optimized for its data, using the delay distribution obtained with the rubber ball.
This policy doesn't improve over the Baseline, suggesting that the uncertainties in the release are also affected by the object's characteristics, thus the delay should be estimated separately for each object.

In \cref{table_opt_delays} (right) we report the delay distribution parameters estimated by the optimization procedure in \Cref{sec:delay_dist_opt} on all three objects: the rubber ball, the tennis ball, and the cube.

\begin{table}[t]
\caption{MC-PILOT main parameters for simulation and real setup}
\label{table_hyperparam}
\begin{center}
\begin{tabular}{c||ccc}
\hline
Parameter & Simulation & Real & \\
\hline
$N_{exp}$ & 5 & 10 & \\
$N_a$ & 0 & 2 & \\
$N_{opt}$ & 1500 & 1500 & \\
$M$ & 400 & 400 & \\
$M_d$ & 10 & 10 & \\
$N_b$ & 250 & 250 & \\
$u_M$ & 3.5 & 2.8 & [$m/s$]\\
$T_s$ & 0.01 & 1/60 & [$s$]\\
$T$ & 1.0 & 1.0 & [$s$]\\
$\ell_c$ & 0.1 & 0.1 & [$m$]\\
$\Sigma_{\pi}$ & $\frac{1}{2}I$ & $\frac{1}{2}I$ \\
$\ell_m$ & 0.75 & 0.7 & [$m$]\\
$\ell_M$ & 2.4 & 1.75 & [$m$] \\
$\gamma_M$ & $\frac{\pi}{6}$ & $\frac{\pi}{6}$ & [rad] \\
\hline
\end{tabular}
\end{center}
\end{table}

\begin{table}[t]
\caption{Optimized delay distributions for each object}
\label{table_opt_delays}
\centering
\begin{tabular}{|c|c|c||c|c|c|}
\hline
\multicolumn{3}{|c||}{Simulation (mean $\pm$ std)} & \multicolumn{3}{c|}{Real setup} \\
\hline
\hline
Object & a $[s]$ & b $[s]$ & Object & a $[s]$ & b $[s]$\\
\hline
ball & $0.021$ & $0.003 $ & rubber ball & 0.259 & 0.009 \\
 & $\pm 0.057$ & $\pm 0.004$ & tennis ball & 0.258 & 0.008 \\
 &  &  & cube & 0.265 & 0.002 \\

\hline
\end{tabular}    
\end{table}

\begin{figure}
    \centering
    \subfloat{\includegraphics[width=0.5\columnwidth]{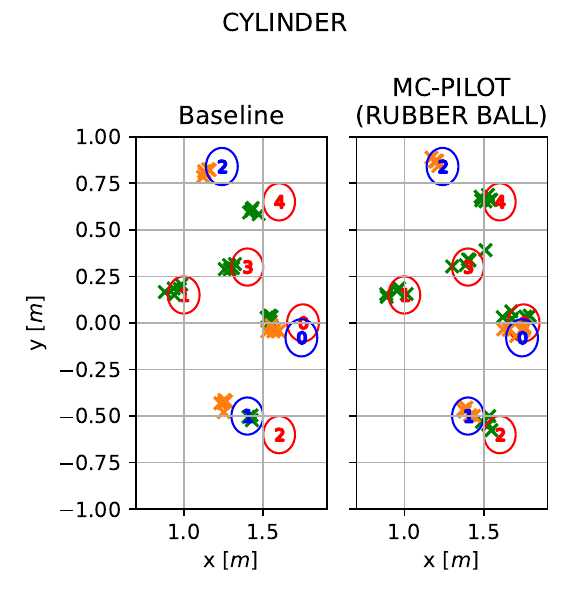}}
    \subfloat{\includegraphics[width=0.5\columnwidth]{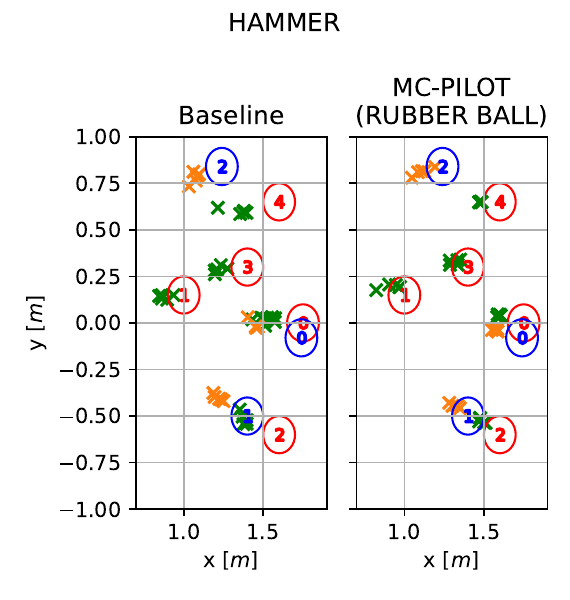}}

    \subfloat{\includegraphics[width=0.5\columnwidth]{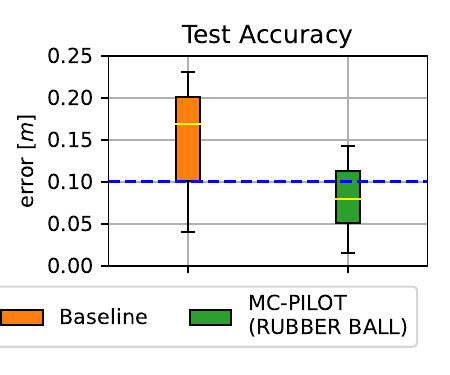}}  
    \subfloat{\includegraphics[width=0.5\columnwidth]{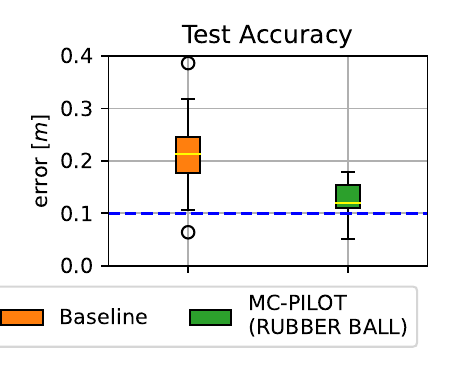}}
    \caption{Accuracy tests with the cylinder (left) and the hammer (right). Each subplot shows the performance of a policy. The numbered red circles are placed in the test positions, while the blue circles are placed in the demo positions, their radius is $\SI{0.1}{\m}$.}
    \label{fig:results_cylinder_hammer}
\end{figure}

Finally, we show the results of the policy trained with MC-PILOT on the rubber ball, on two more objects, a cylinder and a hammer in \cref{fig:results_cylinder_hammer}. Indeed, the free fall dynamics of these two objects cannot be described by the motion of their geometrical center, as the moment of inertia has remarkable effects on their motion. This means that the motion depends on the orientation of the object, since the vision system does not record this data, it is not presented to the GP model, therefore it cannot learn its dynamics. Still, as with the other tested objects, the policy trained on the rubber ball performs more accurately than the baseline.

\emph{Changing task requirements.}
When task requirements change in PnT tasks, learning-based algorithms typically require re-exploration of the environment to adapt the policy to the new task objectives. Changes can include, in general, a modification of the targets' domain, described in \Cref{eq:domain}, for example, a change in height. Exploiting the Model-Based framework, MC-PILOT can adapt the throwing policy to the new task specifications without performing a new exploration. Namely, it only requires re-executing the policy optimization step with the updated requirements. This step requires about $15$ minutes on a consumer Nvidia RTX 3060 laptop GPU. In a demonstrative experiment, we consider a bin target instead of spacial ground targets, the target domain $\D_P$ is updated, changing $z_p$ to match the top of the bin. The bin, shown in \Cref{fig:robot_pic}, is manually moved to arbitrary locations within its domain.
In the demonstration video \footnote{ \url{https://youtu.be/0e8IWstunsc}} we show the qualitative performance of MC-PILOT for the object-throwing task with the target bin, considering the rubber ball, cube, and tennis ball objects. The experiment is also executed with the baseline, while we do not evaluate the Neural Network policy, which instead would require recollecting training data.
The policy trained by MC-PILOT reaches 100\% target reach accuracy, with all three tested objects, while the Baseline struggles with the furthest targets, reaching an accuracy of $\sim$50\%.

\section{Conclusions}
\label{sec:conclusions}
This paper presents a learning-based framework to solve PnT tasks: MC-PILOT. MC-PILOT is a Model-Based Reinforcement Learning algorithm with high sample efficiency, able to learn a throwing policy for industrial manipulators from a few data samples.
MC-PILOT learns a model of the system dynamics from collected data through GPR, including a distribution of the gripper's opening delay. The control policy is optimized on the learned model employing a particle-based method, which takes into account release errors due to the gripper actuation delay, as well as model uncertainties.

We demonstrate the effectiveness of the proposed algorithm on a simulated and real setup, comparing it with an analytical and a Model-Free RL solution.
Results confirm the advantages of the MBRL approach concerning data efficiency, highlighting the importance of modeling communication and control errors in the policy optimization algorithm.
Compared to analytical solutions, MC-PILOT allows to reach near-perfect task accuracy, while requiring much less training data, w.r.t. comparable Model-Free learning-based solutions. Moreover, the Model-Based framework allows the adaptation of the policy to updated task requirements, avoiding the collection of new training data.

\section*{Acknowledgements}
The authors wish to thank Riccardo Lorigiola and Nicolas Viezzer from the SPARCS Laboratory of the Department of Information Engineering of the University of Padova. Their valuable contribution to the design and realization of the prosthetic tooltips made this work possible.

Alberto Dalla Libera and Giulio Giacomuzzo  were supported by PNRR research activities of the consortium iNEST (Interconnected North-Est Innovation Ecosystem) funded by the European Union Next GenerationEU (Piano Nazionale di Ripresa e Resilienza (PNRR) – Missione 4 Componente 2, Investimento 1.5 – D.D. 1058  23/06/2022, ECS\_00000043). This manuscript reflects only the Authors’ views and opinions, neither the European Union nor the European Commission can be considered responsible for them.


\begin{thebibliography}{10}
	\expandafter\ifx\csname url\endcsname\relax
	\def\url#1{\texttt{#1}}\fi
	\expandafter\ifx\csname urlprefix\endcsname\relax\def\urlprefix{URL }\fi
	\expandafter\ifx\csname href\endcsname\relax
	\def\href#1#2{#2} \def\path#1{#1}\fi
	
	\bibitem{pnp}
	R.~A. Brooks, \href{https://doi.org/10.1177/027836498300200402}{Planning
		collision- free motions for pick-and-place operations}, The International
	Journal of Robotics Research 2~(4) (1983) 19--44.
	\newblock \href
	{http://arxiv.org/abs/https://doi.org/10.1177/027836498300200402}
	{\path{arXiv:https://doi.org/10.1177/027836498300200402}}, \href
	{https://doi.org/10.1177/027836498300200402}
	{\path{doi:10.1177/027836498300200402}}.
	\newline\urlprefix\url{https://doi.org/10.1177/027836498300200402}
	
	\bibitem{5342467}
	Y.~Zhang, B.~K. Chen, X.~Liu, Y.~Sun, Autonomous robotic pick-and-place of
	microobjects, IEEE Transactions on Robotics 26~(1) (2010) 200--207.
	\newblock \href {https://doi.org/10.1109/TRO.2009.2034831}
	{\path{doi:10.1109/TRO.2009.2034831}}.
	
	\bibitem{extrinsic_dexterity}
	N.~C. Dafle, A.~Rodriguez, R.~Paolini, B.~Tang, S.~S. Srinivasa, M.~Erdmann,
	M.~T. Mason, I.~Lundberg, H.~Staab, T.~Fuhlbrigge, Extrinsic dexterity:
	In-hand manipulation with external forces, in: IEEE ICRA, 2014, pp.
	1578--1585.
	\newblock \href {https://doi.org/10.1109/ICRA.2014.6907062}
	{\path{doi:10.1109/ICRA.2014.6907062}}.
	
	\bibitem{pmlr-v164-chen22a}
	T.~Chen, J.~Xu, P.~Agrawal,
	\href{https://proceedings.mlr.press/v164/chen22a.html}{A system for general
		in-hand object re-orientation}, in: A.~Faust, D.~Hsu, G.~Neumann (Eds.),
	Proceedings of the 5th Conference on Robot Learning, Vol. 164 of Proceedings
	of Machine Learning Research, PMLR, 2022, pp. 297--307.
	\newline\urlprefix\url{https://proceedings.mlr.press/v164/chen22a.html}
	
	\bibitem{urban_waste_throwing}
	F.~Raptopoulos, M.~Koskinopoulou, M.~Maniadakis, Robotic pick-and-toss
	facilitates urban waste sorting, in: 2020 IEEE 16th International Conference
	on Automation Science and Engineering (CASE), 2020, pp. 1149--1154.
	\newblock \href {https://doi.org/10.1109/CASE48305.2020.9216746}
	{\path{doi:10.1109/CASE48305.2020.9216746}}.
	
	\bibitem{lecun2015deep_learning}
	Y.~LeCun, Y.~Bengio, G.~Hinton, Deep learning, nature 521~(7553) (2015)
	436--444.
	
	\bibitem{ccalicsir2019mfrl_survey}
	S.~{\c{C}}al{\i}{\c{s}}{\i}r, M.~K. Pehlivano{\u{g}}lu, Model-free
	reinforcement learning algorithms: A survey, in: 2019 27th signal processing
	and communications applications conference (SIU), IEEE, 2019, pp. 1--4.
	
	\bibitem{RL_policy_training_ball_throwing}
	A.~Ghadirzadeh, A.~Maki, D.~Kragic, M.~Björkman, Deep predictive policy
	training using reinforcement learning, in: 2017 IEEE/RSJ International
	Conference on Intelligent Robots and Systems (IROS), 2017, pp. 2351--2358.
	\newblock \href {https://doi.org/10.1109/IROS.2017.8206046}
	{\path{doi:10.1109/IROS.2017.8206046}}.
	
	\bibitem{hierarchical_RL_ball_throwing}
	R.~Pinsler, R.~Akrour, T.~Osa, J.~Peters, G.~Neumann, Sample and feedback
	efficient hierarchical reinforcement learning from human preferences, in:
	2018 IEEE International Conference on Robotics and Automation (ICRA), 2018,
	pp. 596--601.
	\newblock \href {https://doi.org/10.1109/ICRA.2018.8460907}
	{\path{doi:10.1109/ICRA.2018.8460907}}.
	
	\bibitem{kober2012reinforcement_learning_primitives}
	J.~Kober, A.~Wilhelm, E.~Oztop, J.~Peters, Reinforcement learning to adjust
	parametrized motor primitives to new situations, Autonomous Robots 33 (2012)
	361--379.
	
	\bibitem{gutzeit2018besman_ball_throwing}
	L.~Gutzeit, A.~Fabisch, M.~Otto, J.~H. Metzen, J.~Hansen, F.~Kirchner, E.~A.
	Kirchner, The besman learning platform for automated robot skill learning,
	Frontiers in Robotics and AI 5 (2018) 43.
	
	\bibitem{robot_skill_learning_deep_autoencoder_ball_throwing}
	R.~Pahič, Z.~Lončarević, A.~Gams, A.~Ude, Robot skill learning in latent
	space of a deep autoencoder neural network, Robotics and Autonomous Systems
	135 (2021) 103690.
	\newblock \href {https://doi.org/https://doi.org/10.1016/j.robot.2020.103690}
	{\path{doi:https://doi.org/10.1016/j.robot.2020.103690}}.
	
	\bibitem{ball_throwing_visual_feedbacK_gripper}
	J.-S. Hu, M.-C. Chien, Y.-J. Chang, S.-H. Su, C.-Y. Kai, A ball-throwing robot
	with visual feedback, in: 2010 IEEE/RSJ International Conference on
	Intelligent Robots and Systems, 2010, pp. 2511--2512.
	\newblock \href {https://doi.org/10.1109/IROS.2010.5649335}
	{\path{doi:10.1109/IROS.2010.5649335}}.
	
	\bibitem{learning_ball_throwing_decision_transformers_gripper}
	M.~Monastirsky, O.~Azulay, A.~Sintov, Learning to throw with a handful of
	samples using decision transformers, IEEE Robotics and Automation Letters
	8~(2) (2023) 576--583.
	\newblock \href {https://doi.org/10.1109/LRA.2022.3229266}
	{\path{doi:10.1109/LRA.2022.3229266}}.
	
	\bibitem{huang2023dynamic_handover_gripper}
	B.~Huang, Y.~Chen, T.~Wang, Y.~Qin, Y.~Yang, N.~Atanasov, X.~Wang, Dynamic
	handover: Throw and catch with bimanual hands, in: Conference on Robot
	Learning, PMLR, 2023, pp. 1887--1902.
	
	\bibitem{tossingbot}
	A.~Zeng, S.~Song, J.~Lee, A.~Rodriguez, T.~Funkhouser, Tossingbot: Learning to
	throw arbitrary objects with residual physics, IEEE Transactions on Robotics
	36~(4) (2020) 1307--1319.
	\newblock \href {https://doi.org/10.1109/TRO.2020.2988642}
	{\path{doi:10.1109/TRO.2020.2988642}}.
	
	\bibitem{pick_and_throw_sorting_deep_rl}
	Z.~Fang, Y.~Hou, J.~Li, A pick-and-throw method for enhancing robotic sorting
	ability via deep reinforcement learning, in: 2021 36th Youth Academic Annual
	Conference of Chinese Association of Automation (YAC), 2021, pp. 479--484.
	\newblock \href {https://doi.org/10.1109/YAC53711.2021.9486466}
	{\path{doi:10.1109/YAC53711.2021.9486466}}.
	
	\bibitem{moerland2023mbrl_survey}
	T.~M. Moerland, J.~Broekens, A.~Plaat, C.~M. Jonker, et~al., Model-based
	reinforcement learning: A survey, Foundations and Trends{\textregistered} in
	Machine Learning 16~(1) (2023) 1--118.
	
	\bibitem{mcpilco_tro}
	F.~Amadio, A.~Dalla~Libera, R.~Antonello, D.~Nikovski, R.~Carli, D.~Romeres,
	Model-based policy search using monte carlo gradient estimation with real
	systems application, IEEE Transactions on Robotics 38~(6) (2022) 3879--3898.
	\newblock \href {https://doi.org/10.1109/TRO.2022.3184837}
	{\path{doi:10.1109/TRO.2022.3184837}}.
	
	\bibitem{amadio2023mcpilco_raw_meas}
	F.~Amadio, A.~Dalla~Libera, D.~Nikovski, R.~Carli, D.~Romeres, Learning control
	from raw position measurements, in: 2023 American Control Conference (ACC),
	IEEE, 2023, pp. 2171--2178.
	
	\bibitem{mcpilco_robot_tossing_sim}
	N.~Turcato, A.~Dalla~Libera, G.~Giacomuzzo, R.~Carli, Teaching a robot to toss
	arbitrary objects with model-based reinforcement learning, in: 2023 9th
	International Conference on Control, Decision and Information Technologies
	(CoDIT), 2023, pp. 1126--1131.
	\newblock \href {https://doi.org/10.1109/CoDIT58514.2023.10284290}
	{\path{doi:10.1109/CoDIT58514.2023.10284290}}.
	
	\bibitem{wiebe2024reinforcement}
	F.~Wiebe, N.~Turcato, A.~Dalla~Libera, C.~Zhang, T.~Vincent, S.~Vyas,
	G.~Giacomuzzo, R.~Carli, D.~Romeres, A.~Sathuluri, et~al., Reinforcement
	learning for athletic intelligence: Lessons from the 1st “ai olympics with
	realaigym” competition,”, in: Proceedings of the Thirty-Third
	International Joint Conference on Artificial Intelligence, IJCAI-24 (K.
	Larson, ed.), 2024, pp. 8833--8837.
	
	\bibitem{taylor2019optimal_motion_planning}
	O.~Taylor, A.~Rodriguez, Optimal shape and motion planning for dynamic planar
	manipulation, Autonomous Robots 43 (2019) 327--344.
	
	\bibitem{stochastic_motion_planning_obj_throwing}
	A.~Sintov, A.~Shapiro, A stochastic dynamic motion planning algorithm for
	object-throwing, in: 2015 IEEE International Conference on Robotics and
	Automation (ICRA), 2015, pp. 2475--2480.
	\newblock \href {https://doi.org/10.1109/ICRA.2015.7139530}
	{\path{doi:10.1109/ICRA.2015.7139530}}.
	
	\bibitem{paraschos2013probabilistic_primitives_robotics_jan_peters}
	A.~Paraschos, C.~Daniel, J.~R. Peters, G.~Neumann, Probabilistic movement
	primitives, Advances in neural information processing systems 26 (2013).
	
	\bibitem{DQN}
	V.~Mnih, K.~Kavukcuoglu, D.~Silver, A.~A. Rusu, J.~Veness, M.~G. Bellemare,
	A.~Graves, M.~Riedmiller, A.~K. Fidjeland, G.~Ostrovski, et~al., Human-level
	control through deep reinforcement learning, nature 518~(7540) (2015)
	529--533.
	
	\bibitem{DDPG}
	T.~P. Lillicrap, J.~J. Hunt, A.~Pritzel, N.~Heess, T.~Erez, Y.~Tassa,
	D.~Silver, D.~Wierstra, Continuous control with deep reinforcement learning,
	arXiv preprint arXiv:1509.02971 (2015).
	
	\bibitem{softToss_gripper}
	D.~Bianchi, M.~G. Antonelli, C.~Laschi, A.~M. Sabatini, E.~Falotico, Softoss:
	Learning to throw objects with a soft robot, IEEE Robotics \& Automation
	Magazine (2023) 2--12\href {https://doi.org/10.1109/MRA.2023.3310865}
	{\path{doi:10.1109/MRA.2023.3310865}}.
	
	\bibitem{della2020soft_robots}
	C.~Della~Santina, M.~G. Catalano, A.~Bicchi, M.~Ang, O.~Khatib, B.~Siciliano,
	Soft robots, Encyclopedia of Robotics 489 (2020).
	
	\bibitem{ppo}
	J.~Schulman, F.~Wolski, P.~Dhariwal, A.~Radford, O.~Klimov, Proximal policy
	optimization algorithms, arXiv preprint arXiv:1707.06347 (2017).
	
	\bibitem{DT}
	M.~Monastirsky, O.~Azulay, A.~Sintov, Learning to throw with a handful of
	samples using decision transformers, IEEE Robotics and Automation Letters
	8~(2) (2022) 576--583.
	
	\bibitem{rasmussen2003gps_for_ml}
	C.~E. Rasmussen, Gaussian processes in machine learning, in: Summer school on
	machine learning, Springer, 2003, pp. 63--71.
	
	\bibitem{kernel_methods_and_gp_control_systems_magazine}
	A.~Carè, R.~Carli, A.~D. Libera, D.~Romeres, G.~Pillonetto, Kernel methods and
	gaussian processes for system identification and control: A road map on
	regularized kernel-based learning for control, IEEE Control Systems Magazine
	43~(5) (2023) 69--110.
	\newblock \href {https://doi.org/10.1109/MCS.2023.3291625}
	{\path{doi:10.1109/MCS.2023.3291625}}.
	
	\bibitem{caflisch1998monte_carlo_sampling_ref}
	R.~E. Caflisch, Monte carlo and quasi-monte carlo methods, Acta numerica 7
	(1998) 1--49.
	
	\bibitem{bottou2010large_scale_learning_sgd}
	L.~Bottou, Large-scale machine learning with stochastic gradient descent, in:
	Proc of COMPSTAT'2010, Springer, 2010, pp. 177--186.
	
	\bibitem{kingma2013reparametrization_trick}
	D.~P. Kingma, M.~Welling, Auto-encoding variational bayes, arXiv preprint
	arXiv:1312.6114 (2013).
	
	\bibitem{kingma2014adam}
	D.~P. Kingma, J.~Ba, Adam: A method for stochastic optimization, arXiv preprint
	arXiv:1412.6980 (2014).
	
	\bibitem{srivastava2014dropout}
	N.~Srivastava, G.~Hinton, A.~Krizhevsky, I.~Sutskever, R.~Salakhutdinov,
	Dropout: a simple way to prevent neural networks from overfitting, JMLR
	15~(1) (2014) 1929--1958.
	
	\bibitem{frazier2018tutorial_bayesian_opt}
	P.~I. Frazier, A tutorial on bayesian optimization, arXiv preprint
	arXiv:1807.02811 (2018).
	
	\bibitem{review_bayesian_optimization}
	B.~Shahriari, K.~Swersky, Z.~Wang, R.~P. Adams, N.~de~Freitas, Taking the human
	out of the loop: A review of bayesian optimization, Proceedings of the IEEE
	104~(1) (2016) 148--175.
	\newblock \href {https://doi.org/10.1109/JPROC.2015.2494218}
	{\path{doi:10.1109/JPROC.2015.2494218}}.
	
	\bibitem{calibration_1}
	D.~Evangelista, D.~Allegro, M.~Terreran, A.~Pretto, S.~Ghidoni, An unified
	iterative hand-eye calibration method for eye-on-base and eye-in-hand setups,
	in: 2022 IEEE 27th International Conference on Emerging Technologies and
	Factory Automation (ETFA), 2022, pp. 1--7.
	\newblock \href {https://doi.org/10.1109/ETFA52439.2022.9921738}
	{\path{doi:10.1109/ETFA52439.2022.9921738}}.
	
	\bibitem{calibration_2}
	D.~Allegro, M.~Terreran, S.~Ghidoni, Multi-camera hand-eye calibration for
	human-robot collaboration in industrial robotic workcells (2024).
	\newblock \href {http://arxiv.org/abs/2406.11392} {\path{arXiv:2406.11392}}.
	
	\bibitem{pytorch}
	A.~Paszke, S.~Gross, S.~Chintala, G.~Chanan, E.~Yang, Z.~DeVito, Z.~Lin,
	A.~Desmaison, L.~Antiga, A.~Lerer, Automatic differentiation in pytorch
	(2017).
	
	\bibitem{bayesian_opt_library}
	F.~Nogueira,
	\href{https://github.com/bayesian-optimization/BayesianOptimization}{{Bayesian
			Optimization}: Open source constrained global optimization tool for {Python}}
	(2014--).
	\newline\urlprefix\url{https://github.com/bayesian-optimization/BayesianOptimization}
	
	\bibitem{koenig2004design_gazebo}
	N.~Koenig, A.~Howard, Design and use paradigms for gazebo, an open-source
	multi-robot simulator, in: 2004 IEEE/RSJ IROS, Vol.~3, IEEE, 2004, pp.
	2149--2154.
	
	\bibitem{koubaa2017robot_operating_system_ros}
	A.~Koub{\^a}a, et~al., Robot Operating System (ROS)., Vol.~1, Springer, 2017.
	
	\bibitem{ALMEDEIJ2008218_drag}
	J.~Almedeij, Drag coefficient of flow around a sphere: Matching asymptotically
	the wide trend, Powder Technology 186~(3) (2008) 218--223.
	
	\bibitem{coleman2014reducing}
	D.~Coleman, I.~Sucan, S.~Chitta, N.~Correll, Reducing the barrier to entry of
	complex robotic software: a moveit! case study, arXiv preprint
	arXiv:1404.3785 (2014).
	
	\bibitem{ros_control}
	S.~Chitta, E.~Marder-Eppstein, W.~Meeussen, V.~Pradeep,
	A.~Rodr{\'i}guez~Tsouroukdissian, J.~Bohren, D.~Coleman, B.~Magyar,
	G.~Raiola, M.~L{\"u}dtke, E.~Fern{\'a}ndez~Perdomo, ros\_control: A generic
	and simple control framework for ros, J. Open Source Softw. (2017).
	
	\bibitem{apriltags}
	E.~Olson, Apriltag: A robust and flexible visual fiducial system, in: 2011 IEEE
	International Conference on Robotics and Automation, 2011, pp. 3400--3407.
	\newblock \href {https://doi.org/10.1109/ICRA.2011.5979561}
	{\path{doi:10.1109/ICRA.2011.5979561}}.
	
\end{thebibliography}
\end{document}